\documentclass[conference]{IEEEtran}
\usepackage{tikz}
\usepackage{amsthm}
\usepackage{algorithm}
\usepackage{algpseudocode}
\usepackage{amsmath}
\usepackage{amssymb}
\usepackage[compress,sort,comma,numbers]{natbib}
\usepackage{booktabs}
\usepackage{bm}
\usepackage{color}
\usepackage{etex}
\usepackage{qtree}
\usepackage{graphicx}
\usepackage{multirow}
\usepackage{stmaryrd}
\usepackage{subcaption}
\usepackage{url}
\usepackage{thmtools}
\usepackage{ctable}
\usepackage{balance}
\usepackage[colorlinks=true,
linkcolor=magenta,
citecolor=blue,
urlcolor=black,
]{hyperref}

\newtheorem{theorem}{Theorem}
\newtheorem{lemma}{Lemma}
\theoremstyle{definition}
\newtheorem{defn}{Definition}[section]
\newcommand{\eg}[0]{\textit{e.g.}}

\newcommand{\etc}[0]{\textit{ etc.}}

\newcommand{\bsym}{\boldsymbol}
\newcommand{\mb}{\mathbf}
\newcommand{\mc}{\mathcal}

\DeclareMathOperator*{\mmin}{min} 

\newcommand*\circled[1]{\tikz[baseline=(char.base)]{
            \node[shape=circle,draw,inner sep=0.5pt] (char) {#1};}}
\begin{document}

\title{TreeGAN: Syntax-Aware Sequence Generation with \\ Generative Adversarial
Networks}

\author{
\IEEEauthorblockN{Xinyue Liu}
\IEEEauthorblockA{Worcester Polytechnic Institute\\
xliu4@wpi.edu}
\and
\IEEEauthorblockN{Xiangnan Kong}
\IEEEauthorblockA{Worcester Polytechnic Institute\\
xkong@wpi.edu}
\and
\IEEEauthorblockN{Lei Liu}
\IEEEauthorblockA{Apple\\
magicliulei@gmail.com}\\
\and
\IEEEauthorblockN{Kuorong Chiang}
\IEEEauthorblockA{Huawei\\
kuorong.chiang@gmail.com}
}
\maketitle
\begin{abstract}
    Generative Adversarial Networks (GANs) have shown great capacity on image generation, in which a discriminative model guides the training of a generative model to construct images that resemble real images. 
    Recently, GANs have been extended from generating images to generating sequences (\eg, poems, music and codes). 
    Existing GANs on sequence generation mainly focus on general sequences, which are grammar-free. 
    In many real-world applications, however, we need to generate sequences in a formal language with the constraint of its corresponding grammar.
    For example, to test the performance of a database, one may want to generate a collection of SQL queries, which are not only similar to the queries of real users, but also follow the SQL syntax of the target database.
    Generating such sequences is highly challenging because both the generator and discriminator of GANs need to consider the structure of the sequences and the given grammar in the formal language. To address these issues, 
    we study the problem of syntax-aware sequence generation with GANs, in which a collection of real sequences and a set of pre-defined grammatical rules are given to both discriminator and generator.
    We propose a novel GAN framework, namely TreeGAN, to incorporate a given Context-Free Grammar (CFG) into the sequence generation process. 
    In TreeGAN, the generator employs a recurrent neural network (RNN) to construct a parse tree.
    Each generated parse tree can then be translated to a valid sequence of the given grammar.
    The discriminator uses a tree-structured RNN to distinguish the generated trees from real trees.
    We show that TreeGAN can generate sequences for any CFG and its generation fully conforms with the given syntax. 
    Experiments on synthetic and real data sets demonstrated that TreeGAN significantly improves the quality of the sequence generation in context-free languages.
\end{abstract}

\section{Introduction}
\label{sec:intro}
Generative Adversarial Network (GAN) is an unsupervised learning framework that consists of a generative network and a discriminative network. We called them the generator ($G$) and the discriminator ($D$) respectively. 
$D$ learns to distinguish whether a data instance is from real world or synthetic. 
$G$ attempts to confuse $D$ by producing high-quality synthetic instances. 
$D$ and $G$ in a GAN framework are trained against each other iteratively until they reach the Nash equilibrium.
A well-trained GAN yields a generator that is capable of producing high quality data instances that look like real ones.

Inspired by the enormous success in image generation and related fields, GANs~\cite{goodfellow2014generative} have recently been extended to sequence generation tasks~\cite{yu2017seqgan, zhang2017adversarial}.
GANs for sequence generation have many important applications in real world.
For instance, in order to build a good query optimizer for a database, researcher may want to generate a large amount of high quality synthetic SQL queries to benchmark the optimizer. 
\begin{figure}[t]
    \centering
    \begin{subfigure}{0.8\columnwidth}
    \includegraphics[width=1.\textwidth]{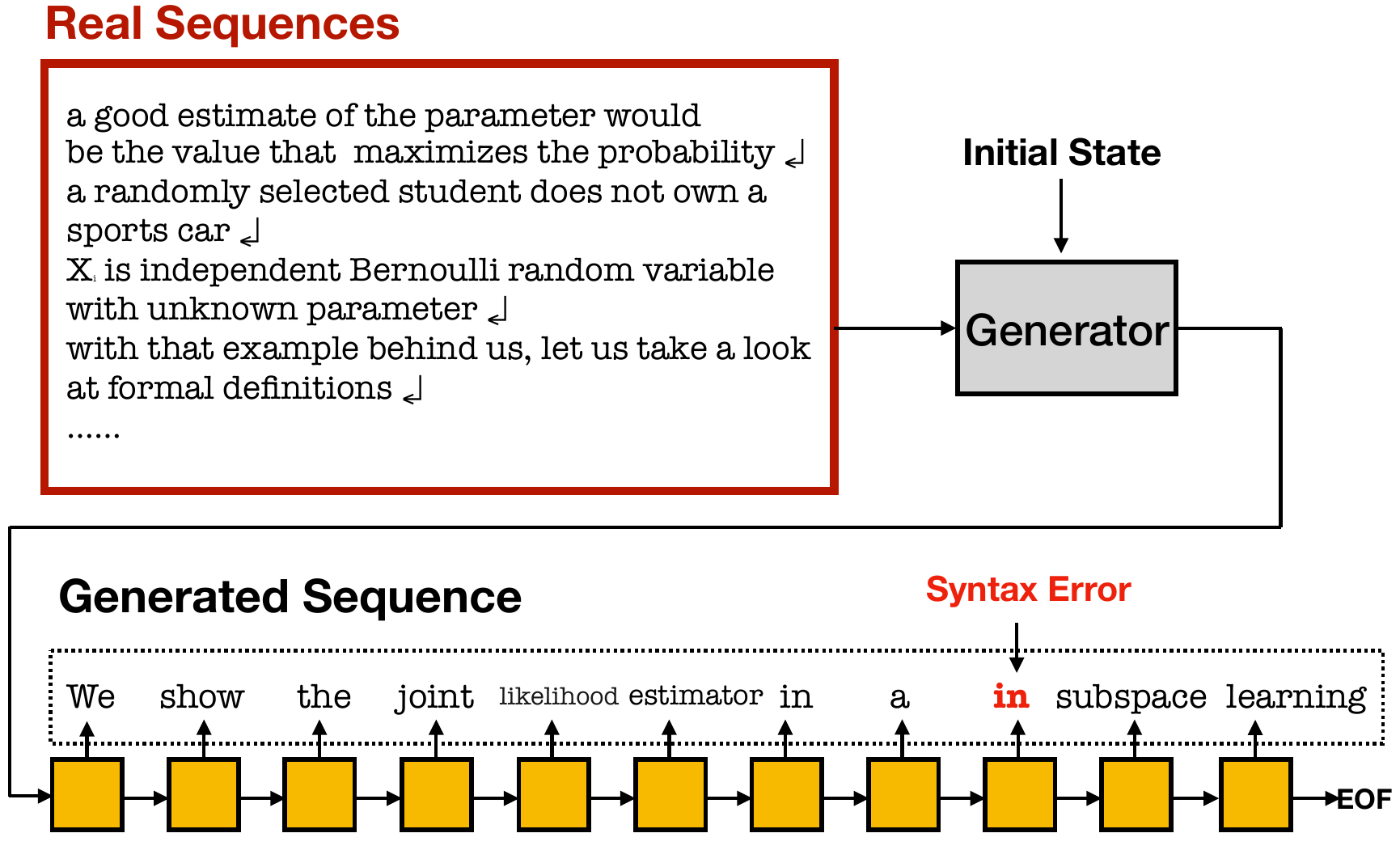}
    \caption{Syntax-Free Sequence Generation Problem (\cite{yu2017seqgan, zhang2017adversarial})}
    \label{fig:conv_prob}
    \end{subfigure}
    \centering
    \begin{subfigure}{0.9\columnwidth}
    \includegraphics[width=1.\textwidth]{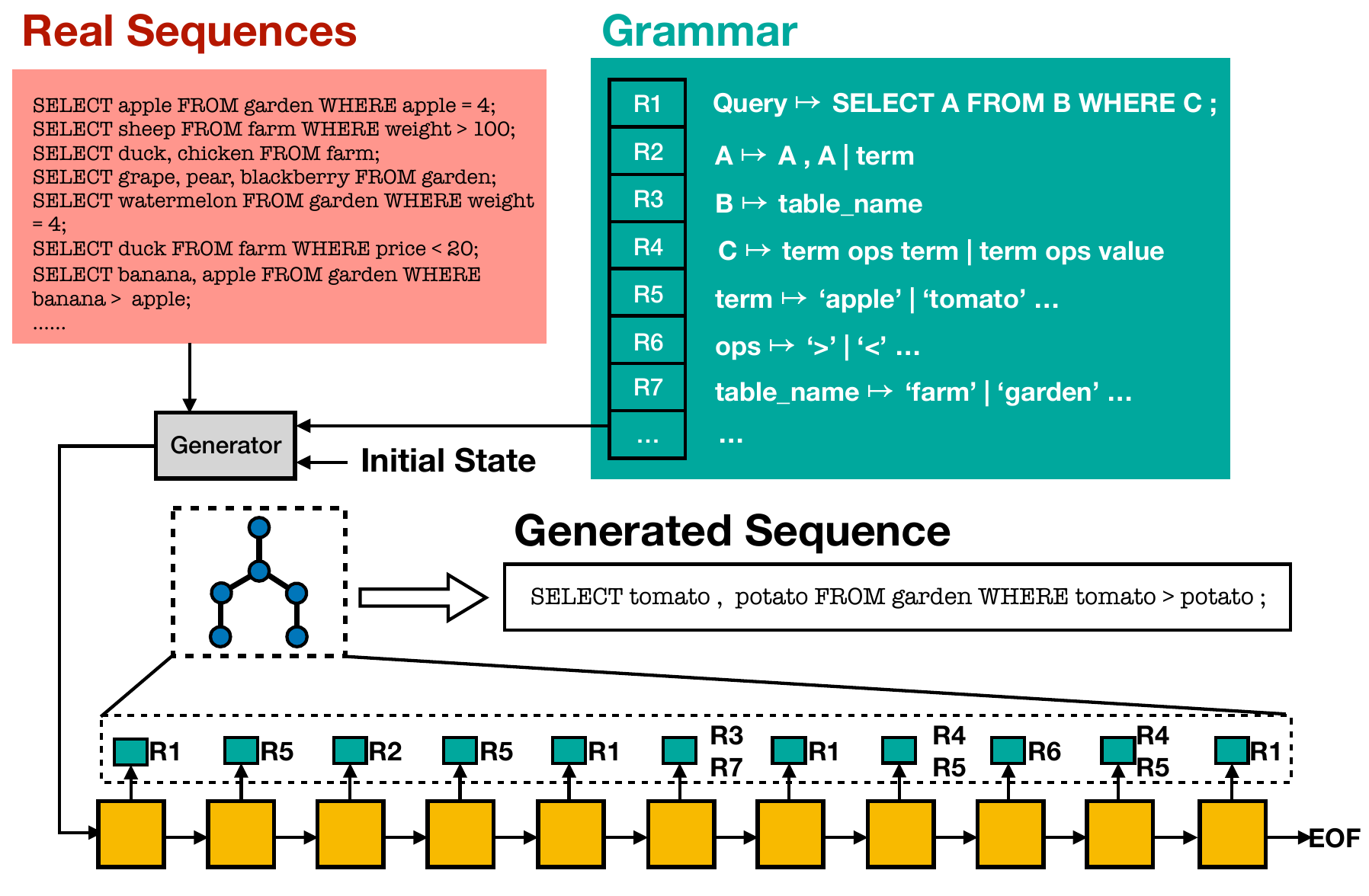}
    \caption{Syntax-Aware Sequence Generation Problem (this paper and \cite{yin2017syntactic})}
    \label{fig:syn_prob}
    \end{subfigure}
    \caption{Comparison of two problem settings. 
    (a) Syntax-free sequence generation problem. Only a set of real sequences are used for training the generator, and the generated sequence may exhibit syntax error.
    (b) Syntax-aware sequence generation problem.
    Besides a set of real sequences (top left box), a set of syntax rules are given as the prior knowledge (top right boxes, {\eg}, ``$A \mapsto A, A$'' and ``$B \mapsto \text{table\_name}$''). 
    At each step, the generator follows one or multiple pre-defined rules (the small boxes in the middle of output arrows, ``R1'',``R2'',\etc) to construct a  sequence (dashed box) that resembles the real sequences and follow the grammar. }
    \vspace{-10pt}
    \label{fig:prob}
\end{figure}
\begin{figure*}[t]
    \centering
    \begin{subfigure}{.25\textwidth}
    \includegraphics[width=1.\linewidth]{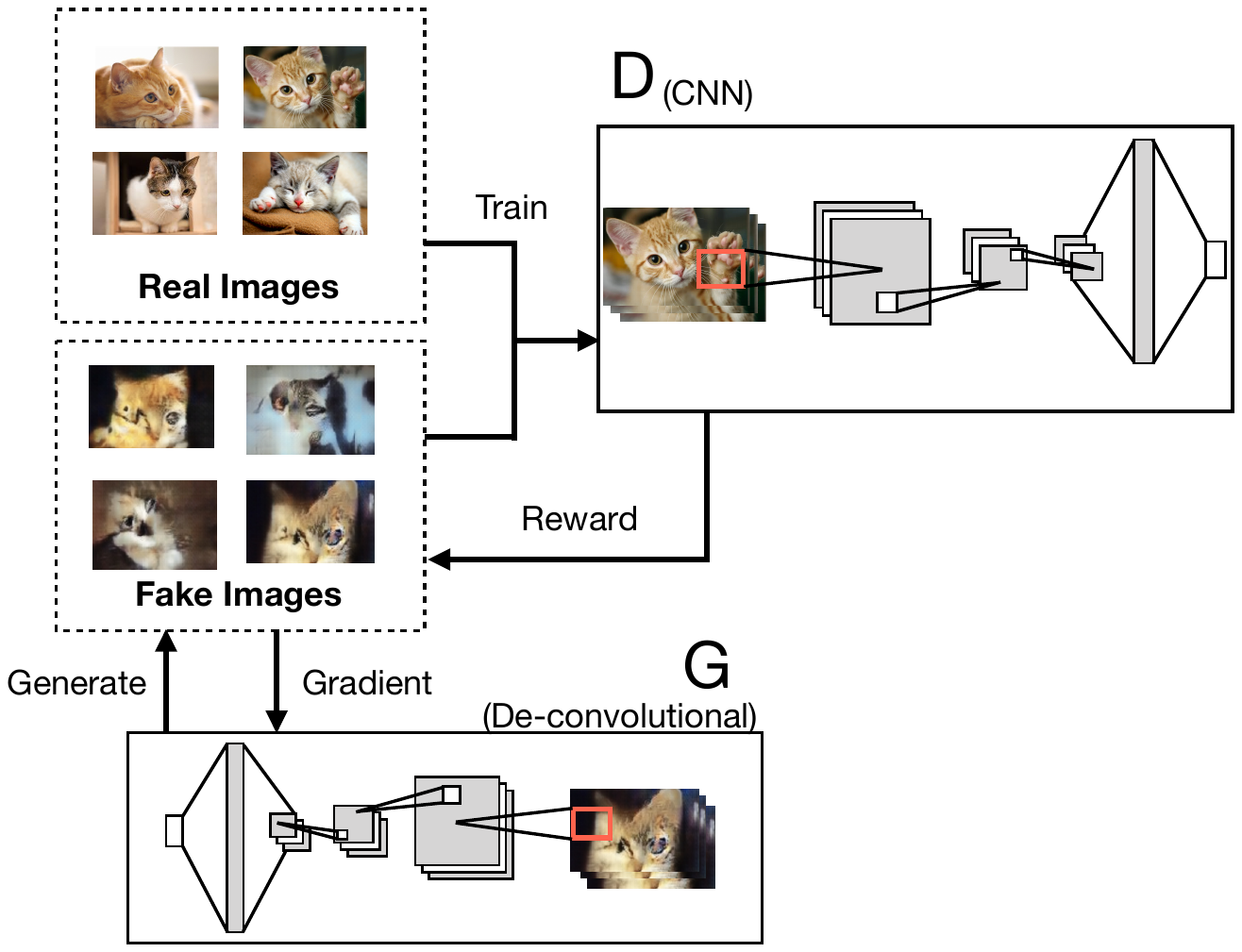}
    \caption{DCGAN for image generation \cite{radford2016unsupervised}} 
    \label{fig:dcgan}
    \end{subfigure}%
    \hspace{5pt}
    \begin{subfigure}{.245\textwidth}
    \centering
    \includegraphics[width=1.\linewidth]{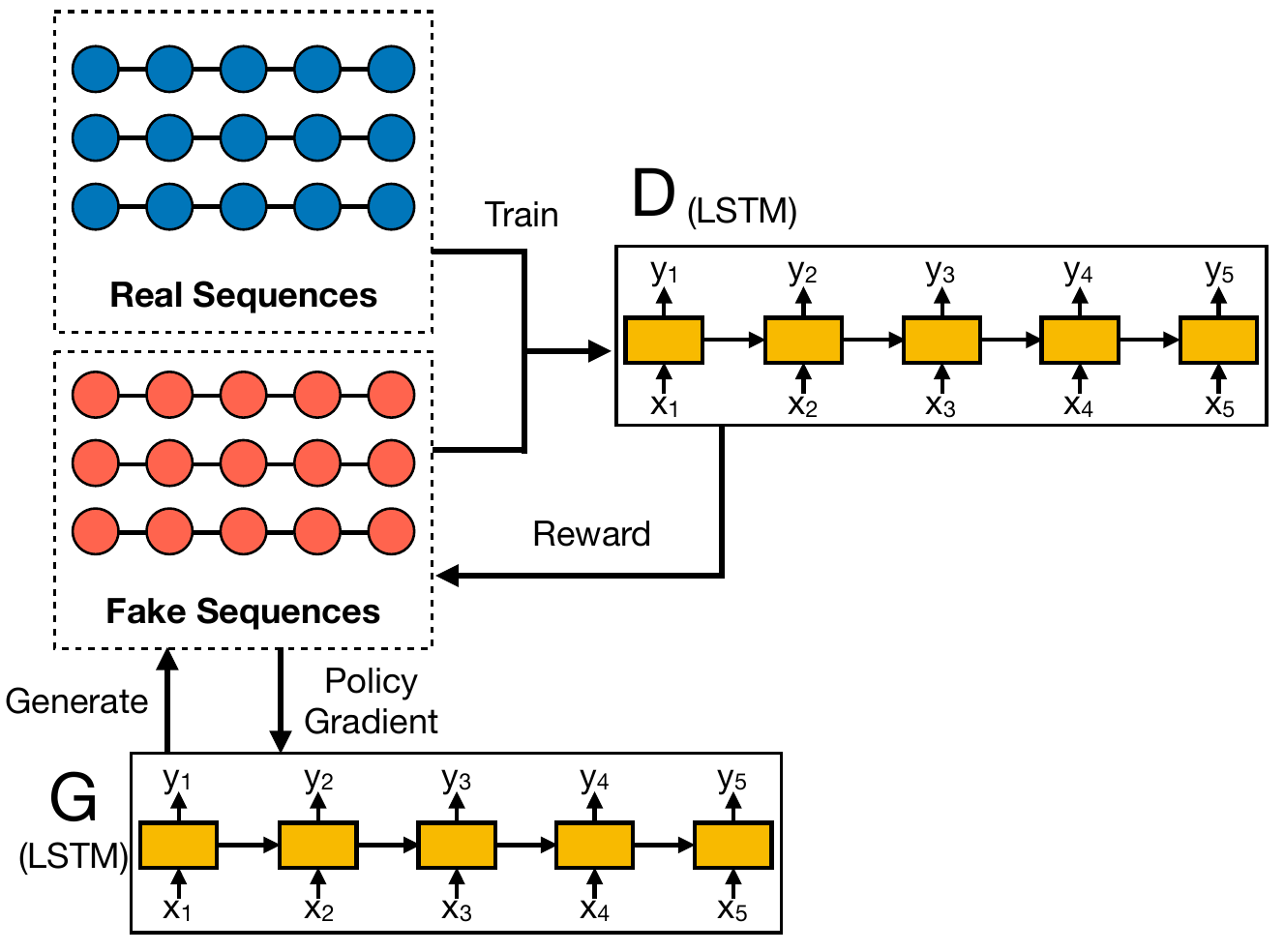}
    \caption{SeqGAN for sequence generation \cite{yu2017seqgan,fedus2018maskgan}}
    \label{fig:seqgan}
    \end{subfigure}
    \hspace{5pt}
    \begin{subfigure}{.40\textwidth}
    \centering
    \includegraphics[width=1.\linewidth]{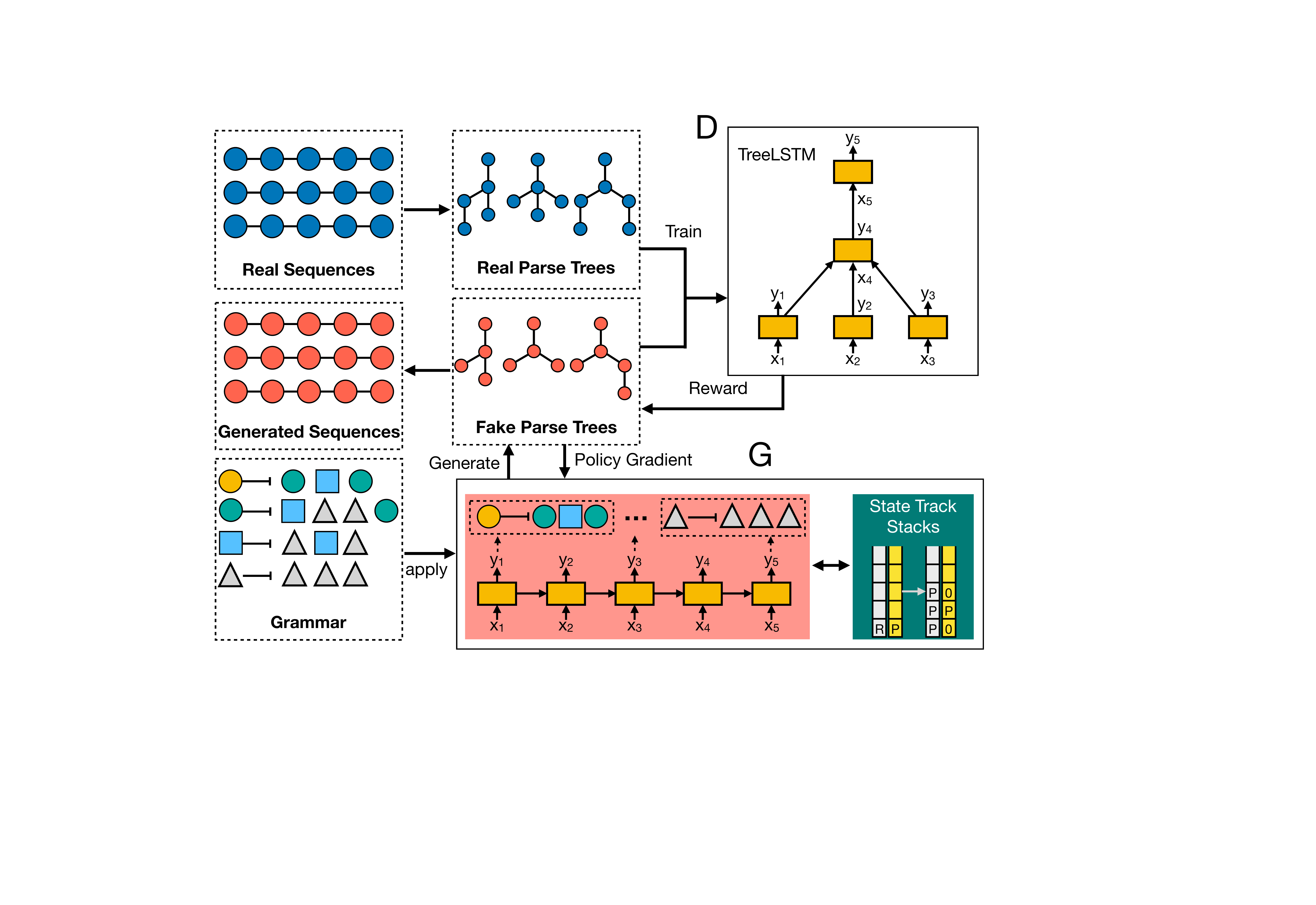}
    \caption{TreeGAN for context-free language generation (this paper)} 
    \label{fig:treegan}
    \end{subfigure}
    \caption{The comparison of related GAN models. ``D" represents the discriminator of the GAN model and ``G" denotes the generator. (a) DCGAN~\cite{radford2016unsupervised}; (b) SeqGAN~\cite{yu2017seqgan} and MaskGAN~\cite{fedus2018maskgan}; (c) TreeGAN (this paper).}
    \vspace{-10pt}
    \label{fig:compare}s
\end{figure*} 
Unlike image generation tasks, most languages have their inherent grammar or syntax.
Existing GAN models ~\cite{yu2017seqgan,zhang2017adversarial,hu2017toward} for sequence generation mainly focus on grammar-free settings as illustrated in Figure~\ref{fig:conv_prob}. 
These methods attempt to learn the complex underlying syntax and grammatical pattern from the data, which is usually highly challenging and requires a large amount of real data samples to achieve a reasonable performance. 
In many formal languages, the grammatical rules or syntax ({\eg},  SQL syntax, Python PL syntax) are pre-defined. 
Incorporating such syntax in GAN training should yield a better sequence generator with syntax-awareness and significantly reduce the searching space during the training phase.
Existing syntax aware sequence generation models \cite{yin2017syntactic} are mainly trained via maximum likelihood estimation (MLE), which highly relies on the quality and quantity of the real data samples.
Some studies \cite{radford2016unsupervised, yu2017seqgan} show that the adversarial training could further improve the generation performance based on MLE.
Even though the existing syntax-aware generation methods incorporate the grammatical information, the generation could be suboptimal.

To tackle above issues, we study the problem of sequence generation under a pre-defined grammar using GANs.
We illustrate this problem setting in Figure.~\ref{fig:syn_prob}, in which a corpus of real sequences (top left box) and a set of grammatical rules (top right box) are given as the input.
The goal is to learn a generative net $G$ that can construct high-quality sequences following the given grammar while resembling the real sequences via adversarial training.
We focus on Context-Free Grammars (CFGs) according to the well-known Chomsky hierarchy~\cite{chomsky1956three}, which can apply to many existing formal languages.
A formal definition of CFGs are provided in Section~\ref{sec:grammar}.
To the best of our knowledge, we make the very first effort to build a syntax aware GAN for sequence generation.

Although GANs have been successfully applied on many tasks, learning such a syntax-aware generative network is not an easy task, which has several challenges:
\begin{itemize}
    \item {\textbf{Guarantee the syntax correctness}}: The difficulty of ensuring the syntax validity lies in the nature of sequence generator: it generates tokens one by one in a sequential order.
    Most syntax models employ a top-down structure like trees to abstract the grammatical information. 
    To fully achieve the syntax awareness, the sequence generator have to follow a certain grammatical tree structure.
    However, the structure of grammatical trees can vary a lot, it is impossible for a sequence generator to cover all the possibilities.  
    \item{\textbf{Tracking the syntax state of incomplete phrase}}: RNN is usually used as the generator in sequence generation, which stores a summary of the generated tokens in its hidden state at each step. 
    However, such summary does not keep track of the syntax information in the partially generated sequence, which leads to possible syntax errors in the entire sequence.
    To build a syntax-aware generator, we need a mechanism that enables RNN to store full syntax information and track the state while generating sequences.  
    \item{\textbf{Syntax-aware discriminator}}: Discriminator is a crucial component of a GAN framework, and should be designed specifically based on the nature of studied task. 
    DCGANs~\cite{radford2016unsupervised} employs Convolutional Neural Networks (CNN)~\cite{krizhevsky2012imagenet} as the discriminator to achieve better performance on image representation and generation, while MaskGAN~\cite{fedus2018maskgan} uses LSTM \cite{hochreiter1997long} as the discriminator to train a sequence generator that fills in missing text. 
    In our problem, simply using LSTM or CNN as the discriminative model could miss critical grammatical pattern, which makes the GAN framework yields weak generator. 
    Hence, a tailored discriminative model should be designed carefully for syntax-aware sequence generation task to encode the rich grammar information of the sequences properly, and to guide the generator to better capture the underlying syntax pattern.
    \item{\textbf{Pre-training}}: 
    A proper pre-training is usually required for both generator and discriminator.
    However, it is unclear that how to design a suitable pre-training strategy for the syntax-aware sequence generation task since we are the first to investigate this problem using GANs.
\end{itemize}

To tackle above challenges, we propose a novel GAN model called TreeGAN. 
Instead of generating sequences directly, TreeGAN absorbs a set of grammatical rules and learns to generate parse trees.
Each generated tree corresponds to a sequence that is valid according to the given grammar.
This approach imposes hard restrictions on the generator, and the syntax correctness of generated sequences is guaranteed. 
We show how these restrictions can be applied in Section \ref{sec:tree_generator}.
Consequently, the vanilla RNN/LSTM is no longer the optimal choice for the discriminator since the generator of TreeGAN is generating trees instead of plain sequences.
To better distinguish the fake parse trees from real parse trees, we use TreeLSTM \cite{TaiSM15} to guide the tree generator during the adversarial training, the details are presented in Sec.~\ref{sec:tree_dis}. The corresponding pre-training strategies are discussed in Sec.~\ref{sec:pretrain}.
The contributions of this work are summarized as follows,
\begin{itemize}
    \item We transform the sequence generation problem into the parse tree generation task to effectively incorporate the structural information. We show that each sequence under a CFG could be translated to a corresponding parse tree, which is used in the proposed TreeGAN to guide the generator producing real look parse trees. 
    \item We propose a tree generator that employs LSTM to generating parse trees that follow a pre-defined context-free grammar. 
    \item We propose an adversarial training framework called TreeGAN, in which a Tree-structured LSTM model \cite{TaiSM15} is used as the discriminator to guide the tree generator constructing parse trees.
    \item Extensive experiments performed on synthetic data sets and real data sets demonstrate that the proposed TreeGAN framework can produce high-quality texts/sequences follow the pre-defined context-free grammar.  
\end{itemize}
The rest of this paper is organized as follows.
We compare our work with the related work in Section~\ref{sec:rel}.
We set the problem formulations in Section~\ref{sec:prob}. 
We show how to solve the proposed problems in Section~\ref{sec:method}.
The experimental results for both synthetic data and real data are shown in Section~\ref{sec:exp}. 
Then we conclude the paper in Section~\ref{sec:con}.

\section{Problem Formulation}
\label{sec:prob}

\begin{table}[t]
    \centering 
    \caption{Summary of Notations.}
    \label{tab:notation}
    \begin{tabular}{|c|l|}
    \hline\vspace{-8pt}&\\
    Symbol& Definition\\
    \hline\vspace{-8pt}&\\
     $\mathbb{G}$ & a context free grammar (CFG)\\
     $\mc{V}$ & the set of non-terminal variables in a CFG \\
     $\mc{T}$ & the set of terminal tokens in a CFG\\
     $\mc{P}$ & the set of production rule in a CFG\\
     $P \in \mc{P}$ & the production rule \\
     $S$ & the start token in a CFG\\
     $G_\theta$ & a generator that parametrized by $\theta$  \\
     $D_\phi$ & a discriminator that parametrized by $\phi$  \\
    \hline \vspace{-5pt}& \\
    $\bsym x$   & the input feature vector \\
    $\bsym h$   & the hidden state vector\\
    $\mathbf W$ & the weight matrix for the input feature\\
    $\mathbf U$ & the weight matrix for the hidden state\\
    $\bsym b$   & the bias vector \\
    $\bsym i$   & the input gate of LSTM \\
    $\bsym f$   & the forget gate of LSTM\\
    $\bsym o$   & the output gate of LSTM\\
    $\bsym u$   & the memory cell of LSTM before input gate\\
    $\bsym c$   & the memory cell of LSTM after input gate\\
    $\Psi $     & the probability output after fully connected layer\\
    $\mathbf M$ & the mask matrix of TreeGAN \\ 
    $\Omega$    &  the stack of TreeGAN\\
    \hline
    \end{tabular}
\end{table}
\subsection{Notation}
Throughout this paper, we use capital alphabet in boldface, {\eg } $\mb{X}$, to denote a matrix, and $x_{ij}$ refers to the entry of $\mb{X}$ at $i$-th row and $j$-th column.
We use lowercase alphabet in boldface, {\eg } $\mb{x}$, to denote a column-based vector, and $x_i$ refers to the $i$-th entry of $\mb{x}$.
We use calligraphic letters to denote sets, {\eg} $\mc{A}, \mc{B}, \mc{C}$.
The important notations used in this paper are summarized in  Table~\ref{tab:notation}.

\subsection{Syntax Aware Sequence Generation}
\label{sec:problem_def}
The syntax-aware sequence generation problem is defined as follows. 
\begin{defn}
Given a dataset of real-world structured sequences $\mc{X} = \{X_1, \dots, X_N\}$, where all $X_n \in \mc{X}$ follows a grammar $\mathbb G$, train a $\theta$-parameterized generative net $G_\theta$ to construct a sequence $Y_{1:T} = (y_1, \dots, y_T)$ with $y_t \in \mc{Y}$, where $\mc{Y}$ is the set of vocabulary of tokens.
\end{defn}

\subsection{Grammar}
\label{sec:grammar}

In this paper, we study the sequence generation problem in context-free grammars (CFGs), which is formulated in the well-known Chomsky hierarchy \cite{chomsky1956three}.
CFGs can apply to many existing formal languages, such as palindrome and SQL.
A CFG is formally defined as $\mathbb G = (\mc{V}, \mc{T}, \mc{P}, S)$, where $\mc{V}$ is a set of non-terminal variables, $\mc{T} = \mc{Y} \cup \{\epsilon \}$ the set of terminal variables\footnote{$\epsilon$ denotes the empty token, alternatively it can be considered as a special symbol that not included in the set of terminal variables.}, $\mc{P}$ the set of production rules, and $S \in \mc{V}$ the start symbol. 
Each production rule $P \in \mc{P}$ follows the form: 
\begin{equation}
\label{eq:p_rule}
\mc{V} \mapsto (\mc{T} \cup \mc{V})^+
\end{equation}

For example, the context free grammar defines palindromes of $0$s and $1$s are $\mathbb{G}_{pal} = (\{P\}, \{0, 1, \epsilon\}, \mc{A}, P)$, where $\mc{A}$ consists of production rules: 
$ \{ P \mapsto\epsilon, P \mapsto0, P \mapsto1, P \mapsto0P0, P \mapsto 1P1$\}.
Accordingly, the palindrome ``$010010$'' could be derived by applying following procedures sequentially:
\begin{align*}
&\textbf{Step 1} &P   &\mapsto 0P0          &[P &\mapsto 0P0] \\
&\textbf{Step 2} &0P0 &\mapsto 01P10        &[P &\mapsto 1P1] \\
&\textbf{Step 3} &01P10 &\mapsto 010P010    &[P &\mapsto 0P0] \\
&\textbf{Step 4} &010P010  &\mapsto 010010  &[P &\mapsto \epsilon]
\end{align*}

\subsection{Parse Tree} 

For each derivation of a CFG sequence, there is a corresponding tree representation called parse tree. 
The parse tree for any sequence follow context free grammar $\mathbb G = (\mc{V}, \mc{T}, \mc{P}, S)$ are trees with following properties:
\begin{enumerate}
	\item The root node is labeled by $S$.
	
	\item The interior node is labeled by a variable in $\mc{V}$.
	
	\item Every leaf is labeled by a terminal in $\mc{T}$.
	
	\item If a node labeled $A$, and its children are labeled $N_1, \dots, N_k$ from left to right. Then $A \mapsto N_1, \dots, N_k$ is a production rule in $\mc{P}$.
\end{enumerate}

\begin{figure}[t]
\centering
\includegraphics[width=0.45\textwidth]{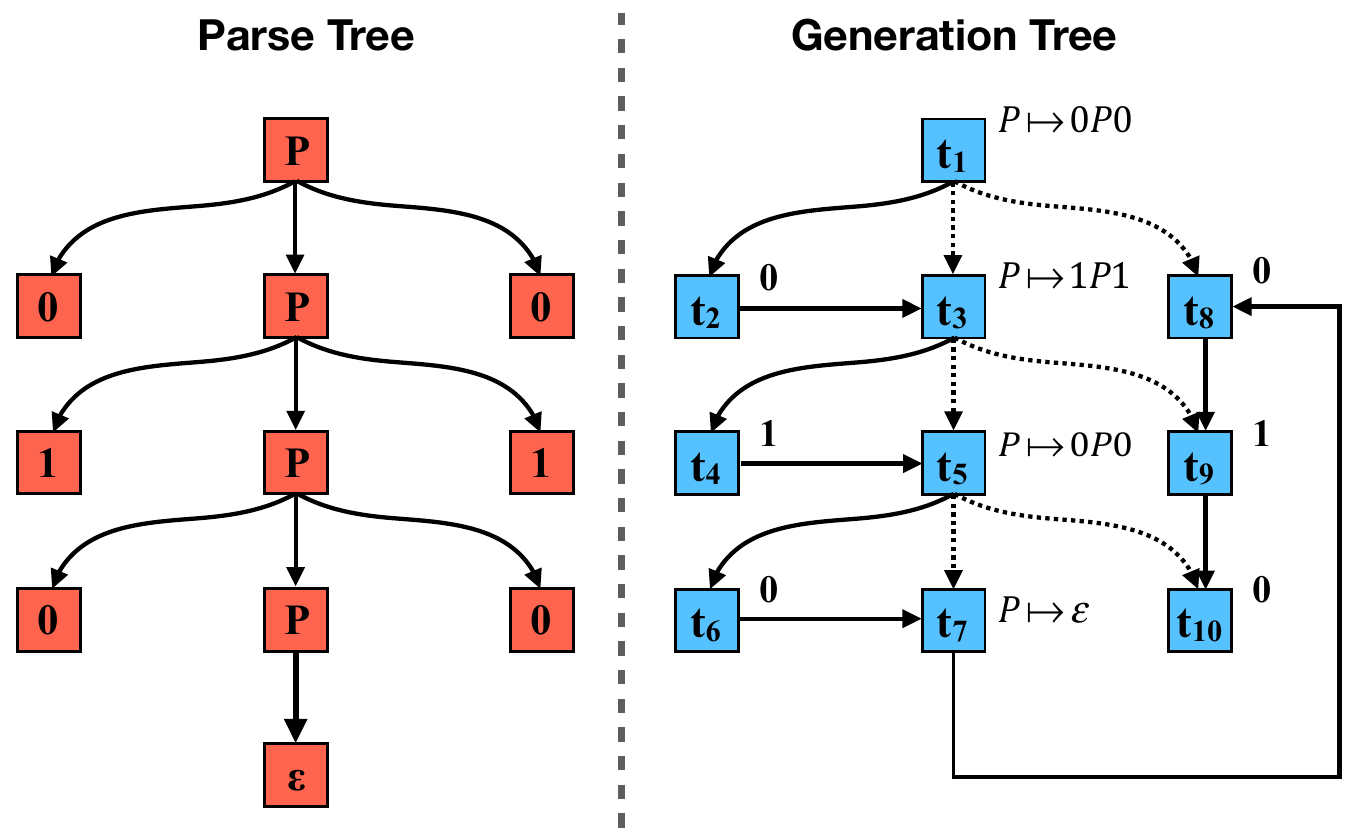}
\vspace{0pt}
\caption{\textbf{Left}: the parse tree for sequence ``010010''. \textbf{Right}: the action sequence used to generate the parse tree shown on the left. The solid arrow denotes the chronological order of the action flow, and the dashed arrow denotes the input of parent embedding (see Sec.~\ref{sec:tree_generator}).} 
\vspace{-10pt}
\label{fig:tree}
\end{figure}

If we concatenate leaves of a parse tree from left to right and top to bottom, we obtain a \textit{yield} of the tree, which is equivalent to the string derived from the root variable.
The parse tree of palindrome sequence $010010$ is illustrated as on L.H.S. in Figure~\ref{fig:tree}.
If we concatenate the leaves of the parse tree shown in Figure~\ref{fig:tree}, we can obtain the sequence ``$010\epsilon010$'', which is equivalent to ``$010010$'' since $\epsilon$ refers to the empty token.

\begin{theorem}
\label{theorem:parse_tree}
	Let $\mathbb G = (\mc{V}, \mc{T}, \mc{P}, S)$ be a CFG. If a sequence $Y$ can be derived using the production rules from $\mc{P}$ and the derivation starts with $S$, then there is a parse tree with root $S$ that yields $Y$. 
\end{theorem}

\begin{proof}
It is equivalent to the proof of Theorem 5.12 in \cite{hopcroft2006introduction}.
\end{proof}

\begin{lemma}
\label{lemma:prod}
If sequence $X$ follows a context free grammar $\mathbb G= (\mc{V}, \mc{T}, \mc{P}, S)$, there is a sequence of productions $Z = (P_1, \dots, P_k)$ that derives $Y$, where $P_1, \dots, P_k \in \mc{P}$. Such mapping can be denoted as $Z \Leftrightarrow X$.
\end{lemma}

\begin{proof}
From Theorem~\ref{theorem:parse_tree} we know there exists a parse tree $Q$ yields $Y$, traversing $Q$ via a depth-first search order yields a sequence of productions $Z$ that derives $Y$.	
\end{proof}

Given Lemma~\ref{lemma:prod}, we can find a set of production sequences $\mc{D} = \{D_1, \dots, D_N \}$ for $\mc{X} = \{X_1, \dots, X_N \}$, where $D_1 \Leftrightarrow X_1, \dots, D_N \Leftrightarrow X_N$.
How to parse each $X_n$ into $D_n$ is out of the scope of this paper and will not be discussed here.

Now we can transform the original syntax-aware sequence generation problem defined in Section~\ref{sec:problem_def} into a parse tree generation problem.

\begin{defn}[Parse Tree Generation Problem]
Given a CFG defined as $\mathbb{G}= (\mc{V}, \mc{T}, \mc{P}, S)$, and $\mc{D} = \{D_1, \dots, D_N \}$ where all the production rules in $\{Z_1, \dots, Z_N\}$ are from $\mc{P}$, the goal is to train a $\theta$-parameterized generative net $G_\theta$ to construct a sequence $Z_{1:T} = (P_1, \dots, P_T)$ with $P_t \in \mc{P}$.
\end{defn}

Additionally, we also train a $\phi$-parameterized discriminative net $D_\phi$ to guide $G_\theta$ to improve the generating quality.
Specifically, $D_\phi(Z)$ is a probability indicating how likely $Z$ is a real data sample.

\section{Methodology}
\label{sec:method}

In this section, we introduce the technical details of our proposed method TreeGAN.
In section \ref{sec:GAN}, We first briefly review the key components of conventional GANs, including its objective function and the optimization approach.
Then we present the detailed design of the tree generator of TreeGAN in section~\ref{sec:tree_generator}, we show how could the generator keep a lossless track information of the syntax state while generating the sequence.
Section~\ref{sec:tree_dis} presents the tailored discriminator we used for TreeGAN.
Finally, in section~\ref{sec:pretrain} we introduce our pre-training strategy, which gives the adversarial training phase a better start point.

\subsection{Generative Adversarial Network}
\label{sec:GAN}
GAN\cite{goodfellow2014generative} aims to obtain the equilibrium of the following optimization objective
\begin{align}
\begin{split}
\label{eq:gan}
\mc{L(\theta, \phi)} = &- \mathbb{E}_{X \sim p_x } \text{log } D_\phi(X) \\
 &- \mathbb{E}_{Y \sim G_\theta} \text{log } (1 - D_\phi(Y))
\end{split}
\end{align}

where $\mc{L}$ is minimized \textit{w.r.t.} $D_\phi$ and is maximized \textit{w.r.t.} $G_\theta$. 
$X$ are sampled from the real-data distribution $p_x$. 
Since the first term of Eq.~(\ref{eq:gan}) does not depend on $G_\theta$,  we only need to consider the second term when training the generator.
However, applying GAN on sequence data has a problem: the gradient of loss from $D_\phi$ \textit{w.r.t} the output of $G_\theta$ is not meaningful for discrete tokens \cite{goodfellow2014generative, yu2017seqgan}.
Thus, we follow the approach proposed in SeqGAN\cite{yu2017seqgan} to use the policy gradient\cite{sutton2000policy} to guide the learning of $G_\theta$.
The reward of $G_\theta$ when given a start state $s_0$ is :
\begin{align}
\begin{split}
\label{eq:reward}
    &\mc{J}(\theta)  = \\
    &\mathbb{E}_{Y \sim G_\theta}
     \text{log}\Big(G_\theta\big(y_1 | s_0\big)
     \prod_{t=2}^T G_\theta\big(y_t|Y_{1:t-1}\big)\Big)
     R(Y_{1:T}),
\end{split}
\end{align}
where $R(\cdot)$ is the reward function for a generated sequence, here we consider the estimated probability of being real by the discriminative net $D_\phi$ as the reward. 
Formally it is defined as
\begin{equation}
    R(Y_{1:T}) = D_\phi\big(Y_{1:T}\big)
\end{equation}

Hence, for sequence generation task, the objective of training the discriminative net is $\text{arg} \mmin_{\phi} \mc{L}(\theta, \phi)$, where $\theta$ is fixed.
And the objective of training the generative net is $\text{arg} \mmin_{\theta} \mc{J}(\theta)$.

\subsection{Tree Generator}
\label{sec:tree_generator}
Inspired by the model proposed in \cite{yin2017syntactic}, we consider the tree generation problem as generating a sequence of actions.
The actions can be categorized into two types, which are (1) the production rules as defined in Eq.~\ref{eq:p_rule} and (2) the terminal tokens in $\mc{V}$.
The R.H.S. of Figure~\ref{fig:tree} illustrates the generation process of the parse tree on L.H.S of Figure~\ref{fig:tree}.
Each node in R.H.S. of Figure~\ref{fig:tree} refers to an action and actions are connected by solid arrows that indict the chronological order of them.
The generation proceeds in depth-first, left-to-right order.
Thus, in order to generate the parse tree shown in Figure~\ref{fig:tree}, the tree generator $G_\theta$ produces the following actions sequentially,
\begin{align*}
P \mapsto 0P0, 0, P \mapsto 1P1, 1, P \mapsto 0P0, 0, P \mapsto \epsilon, 0, 1, 0
\end{align*}

$G_\theta$ starts from the root node at step $t_1$ and proceeds by choosing different production rules to expand the tree, and at leaves, the model generates terminal tokens to close the tree branches.

We employ a vanilla LSTM to implement our tree generator:
\begin{align}
\begin{split}
\label{eq:lstm}
\bsym{i}_t &= \sigma(\mb{W}^{(i)}\bsym{x}_t + \mb{U}^{(i)}\bsym{h}_{t-1} + \bsym{b}^{(i)}), \\
\bsym{f}_t &= \sigma(\mb{W}^{(f)}\bsym{x}_t + \mb{U}^{(f)}\bsym{h}_{t-1} + \bsym{b}^{(f)}), \\
\bsym{o}_t &= \sigma(\mb{W}^{(o)}\bsym{x}_t+ \mb{U}^{(o)}\bsym{h}_{t-1} + \bsym{b}^{(o)}), \\
\bsym{u}_t &= \text{tanh}(\mb{W}^{(u)}\bsym{x}_t + \mb{U}^{(u)}\bsym{h}_{t-1} + \bsym{b}^{(u)}), \\
\bsym{c}_t &= \bsym{i}_t \odot \bsym{u}_t + \bsym{f}_t \odot \bsym{c}_{t-1},\\
\bsym{h}_t &= \bsym{o}_t \odot \text{tanh}(\bsym{c}_t),
\end{split}    
\end{align}
where, $\bsym{i}_t, \bsym{f}_t, \bsym{o}_t, \bsym{c}_t, \bsym{o}_t$ are the input gate, the forget get, the output gate, the memory cell and the hidden state at time step $t$ respectively. 
$\bsym{u}_t$ is the memory cell before input gate at step $t$, and $\odot$ denotes the element-wise multiplication. 

For a data sample $D=(d_1, \dots, d_T)$, the input vector at time step $t$ is $\bsym{x}_t = (\bsym{a}_{t-1}, \bsym{p}_t)$, where $\bsym{a}_{t-1}$ is the action embedding vector for $d_{t-1}$ and $\bsym{p}_t$ is the parent embedding vector for $d_{t}$.

\noindent \textbf{Action Embedding:} 
Two action embedding matrices $\mathbf{W}^{(P)}$ and $\mathbf{W}^{(V)}$ are initialized before train the generator $G_\theta$. Each row in $\mathbf{W}^{(P)}$ ($\mathbf{W}^{(V)}$) corresponds to an embedding vector for an action of production rules (terminal tokens).

\noindent \textbf{Parent Embedding:} 
The tree generator uses the parent feeding illustrated on R.H.S. in Figure~\ref{fig:tree} to inherit the information encoded in the parent action along the generation tree. 
As shown in Figure~\ref{fig:tree}, when generating action at $t_5$, the embedding of its parent action at $t_3$ will be used. 
The parent action step $p(t)$ is formally defined as the time step at which the action node at time step $t$ is initiated.
Specifically, in Figure~\ref{fig:tree}, the action node at time step $t_2, t_3, t_9$ are all initiated at $t_1$ when $G_\theta$ generates the production $P \mapsto 0P0$.
In this case, $p(t_2) = p(t_3) = p(t_9) = t_1$.

\noindent \textbf{Generation State Tracking:}
As we discussed in Section~\ref{sec:intro}, the conventional RNN stores lossy summarization in its hidden state $\bsym{h}$, which only contains incomplete syntax information of the generated part of a sequence. 
For example, at time step $t_2$ in Figure~\ref{fig:tree}, a conventional RNN may generate action $P\mapsto 1P1$ or $P\mapsto0P0$, which violates the pre-defined grammar. 
Thus, we need an extra control on the RNN to track the generation state accurately.
The output of the LSTM at time step $t$ is denoted as $\bsym{o}_t \in \mathbb{R}^{L}$, where $L$ is the size of the set of actions.
At the output layer of each time step, the generator samples an action from the a multinomial distribution denoted by $\text{softmax}(\bsym{o}_t) = (\hat{o}_t^{(1)}, \dots, \hat{o}_t^{(L)}) $, where $\hat{o}_t^{(k)}$ corresponds to the probability of sampling action $a_k$ at time step $t_t$.
A mask matrix $\mathbf{M}^{(G)} \in \{0, 1\}^{(|\mc{V}|\times |\mc{P} \cup \mc{T}| )}$ can be derived for grammar $\mc{G}$.
The $k$-th row in $\mathbf{M}^{(G)}$, which is denoted as $\mathbf{M}^{(G)}(k)$, marks the valid actions for $v_k \in \mc{V}$ as $1$s and the invalid ones as $0$s.
Thus, when the generator $G_\theta$ reaches the time step $t_t$ where the corresponding node is non-terminal node $v_k \in \mc{V}$, then the following masking is performed before it generates the token for step $t_t$:
\begin{equation}
    \tilde{\bsym{o}}_t = \text{softmax}(\bsym{o}_t) \odot \mathbf{M}^{(G)}(k)
\end{equation}
Hence, the probability of invalid actions for $v_k$ is reset to $0$ in $\tilde{\bsym{o}}_t$.
In the other cases, when $G_\theta$ reaches a time step where the corresponding node is a terminal node $y_k \in \mc{T}$, then $y_k$ is directly generated.
By applying such masking process, our tree generator can no longer sample actions that violate the syntax.

\begin{figure}[t]
\begin{minipage}[l]{\columnwidth}
\centering
\includegraphics[width=\textwidth]{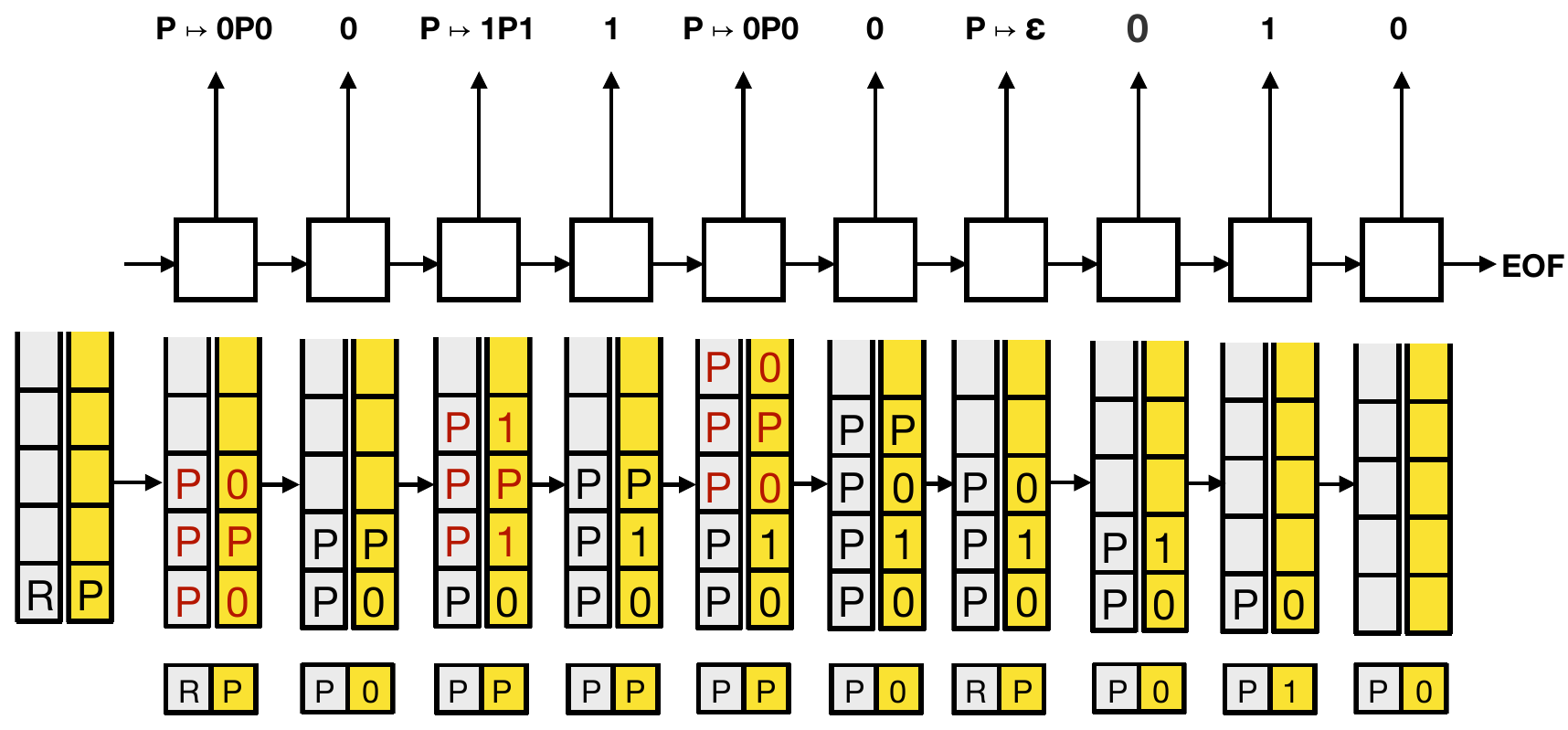}
\end{minipage}
\caption{The generation process of the parse tree shown in Figure~\ref{fig:tree}. The generator maintains a parent stack (grey columns) and a children stack (yellow columns), the current node (yellow boxes below the stacks) and its parent (grey boxes) are popped from the two stacks respectively at each generation step. The red elements in the stack refer to the ones are pushed at each step. The generation terminates when both stacks are empty.}
\label{fig:gen}
\vspace{-10pt}
\end{figure}

\noindent \textbf{Tracking Algorithm:} 
The remaining problem is how the tree generator identifies the node type and retrieves the parent action at time step $t_t$.
As shown in Figure~\ref{fig:gen}, we maintain two stacks $\Omega^{(P)}$ and $\Omega^{(C)}$ for \emph{parent tracking} and \emph{children tracking} respectively, which is analogous to the well-known pushdown automata (PDA).
At the beginning of generation, the stacks are initialized as $\Omega^{(P)}=[\Gamma, R]$ and  $\Omega^{(C)}= [\Gamma, S]$, where $\Gamma$ is the empty stack symbol that cannot be popped and $R$ is the pseudo-root symbol.
At each step $t_t$ of generation, the following stack operations are performed sequentially:
$P  \xleftarrow{\text{pop}}\Omega^{(P)}$,
$C  \xleftarrow{\text{pop}}\Omega^{(C)}$,
where $P$ is the corresponding parent action and $C$ is the head variable for time step $t_t$. 
If $C \in \mc{T}$, then $C$ is generated directly and no further stack operations are required before next time step.

When $C \in \mc{V}$, the embedding of the action at previous time step $t_{t-1}$ and the embedding of $P$ are fetched respectively to build the input vector $\bsym{x}_t = (\bsym{a}_{t-1}, \bsym{p}_t)$. 
After applying Eq.~(\ref{eq:lstm}), an action that takes the form $(C \mapsto \mathtt{H}) \in \mc{P}$ is generated based upon the masked probability vector $\tilde{\bsym o}_t$ , where $\mathtt{H} \in (\mc{V} \cup \mc{T})^+$ is a sequence of variables.
Before moving forward to next time step, the following stack operations are performed, $C \xrightarrow{\text{push}}\Omega^{(P)}$,
     $\textit{reversed}(\mathtt{H}) \xrightarrow{\text{push}}\Omega^{(C)}$,
where we push the variable $C$ into the parent stack, and push the variables in $\mathtt{H}$ into the children stack in a reversed order.

\noindent \textbf{Close A Generation:} If $\Omega^{(P)}=\Omega^{(C)}=[\Gamma]$ at the beginning of a time step, it indicates that all interior nodes have been expanded and all leaves are labeled with a terminal token in the tree, then the generator closes the generation by producing an end symbol.

\subsection{Tree Discriminator}
\label{sec:tree_dis}
Since we require the discriminator encode the rich grammar information of a sequence, it should capture the structure and the semantics of the corresponding parse tree.
Thus, we use the Child-Sum Tree-LSTM~\cite{TaiSM15} as the discriminator of TreeGAN.
The formulation is as follows,
\begin{align}
\begin{split}
\label{eq:treelstm}
\tilde{\bsym{h}}_j & = \sum_{k \in Ch(j)} \bsym{h}_k \\
\bsym{i}_j &= \sigma(\mb{W}^{(i)}\bsym{x}_j + \mb{U}^{(i)}\tilde{\bsym{h}}_j + \bsym{b}^{(i)}), \\
\bsym{f}_{jk} &= \sigma(\mb{W}^{(f)}\bsym{x}_j + \mb{U}^{(f)}\bsym{h}_k + \bsym{b}^{(f)}), \\
\bsym{o}_j &= \sigma(\mb{W}^{(o)}\bsym{x}_j+ \mb{U}^{(o)}\tilde{\bsym{h}}_j + \bsym{b}^{(o)}), \\
\bsym{u}_j &= \text{tanh}(\mb{W}^{(u)}\bsym{x}_j + \mb{U}^{(u)}\tilde{\bsym{h}}_j + \bsym{b}^{(u)}), \\
\bsym{c}_j &= \bsym{i}_j \odot \bsym{u}_j + \sum_{k \in Ch(j)}\bsym{f}_{jk} \odot \bsym{c}_{k},\\
\bsym{h}_j &= \bsym{o}_j \odot \text{tanh}(\bsym{c}_j),
\end{split}
\end{align}
where $Ch(j)$ refers to the set of children of node $j$.
This model is also called Child-Sum Tree-LSTM, in which a tree proceeds from leaves to the root. 
Moreover, $\bsym{h}_r$ denotes the final hidden state for a given tree where $r$ is the root node of the tree, and it encodes the entire tree and can be used for classification.
A fully connected linear layer is appended after the output of Tree-LSTM to obtain the confidence:
\begin{align}
\Psi = \text{sigmoid}(\mb{W}^{(c)}\bsym{h}_r + b^{(c)}),
\end{align}
where $\Psi \in (0, 1)$ refers to the probability of the encoded tree being a real instance.

\subsection{Pre-Training}
\label{sec:pretrain}
Before starting the adversarial training, pre-training of $D_\phi$ and $G_\theta$ are usually required to reach a good initialization, which can facilitate the convergence later in adversarial training.
We initialize the tree generator parameters using conventional maximum likelihood estimation (MLE).
As to the tree discriminator initialization, we let the discriminator distinguish the twisted trees from the real trees.
We randomly swap two subtrees of different head types for each real parse tree in the corpus to construct the twisted tree counterparts. 
The swapping operation breaks the syntax of the real parse tree, which guides the discriminator to learn correct syntax patterns. 

\section{Synthetic Study}
\label{sec:exp}
Due to the lack of well documented syntax and schema (for SQL) in real datasets, we first test the effectiveness of the proposed model on three synthetic datasets with pre-defined syntax and schema as the ground-truth.
In this section, we will first introduce the detailed experimental settings, compared methods and the evaluation metrics used on synthetic study.
We attempt to answer the following research questions within this section.
\begin{itemize}
\item \textbf{RQ1}:	Does TreeGAN correctly capture the syntax information?
\item \textbf{RQ2}: How good does TreeGAN capture the underlying semantical pattern ({\eg } schema)? 
\item \textbf{RQ3}: Could TreeGAN generates sequences of better quliaty when compared to other baselines?
\end{itemize}

\subsection{Dataset}
We prepare three different synthetic datasets with controlled syntax and schema (for SQL datasets only). 

\begin{itemize}
\item \textbf{PLD}: A dataset of palindrome in english alphabet (26 capital letters and 26 lowercase letters).
\item \textbf{SQL-A}: A dataset of SQL queries (\texttt{SELECT} queries) with a small set of grammatical rules.
\item \textbf{SQL-B}: A dataset of SQL queries with larger set of grammatical rules.
\end{itemize}
Note that the proposed TreeGAN uses only the grammar (syntax) but not the schema to train the sequence generator.
The synthetic datasets and the corresponding grammatical rules and schema will be public available after this paper is accepted.

\begin{table}[!h] 
	\small
	\centering
    \caption{Summary of Synthetic Datasets}
    \label{tab:syn_datasets}
    \vspace{0pt}
    \begin{tabular}{|ccccc|}
    \hline
    Dataset   & \# Training  &\# Test       & \# Vocab. & \# Prod. Rules  \\
    \hline
    PLD       & 10,000 & 1,000 & 160    &106  \\
    SQL-A     & 50,000 & 5,000 & 1000   &231 \\
    SQL-B     & 100,000& 5,000 & 5000   &422 \\
    \hline 
    \end{tabular}
    \vspace{0pt}
\end{table}

\begin{figure*}[t]
\centering
\begin{subfigure}{.24\textwidth}
  \includegraphics[width=1.\linewidth]{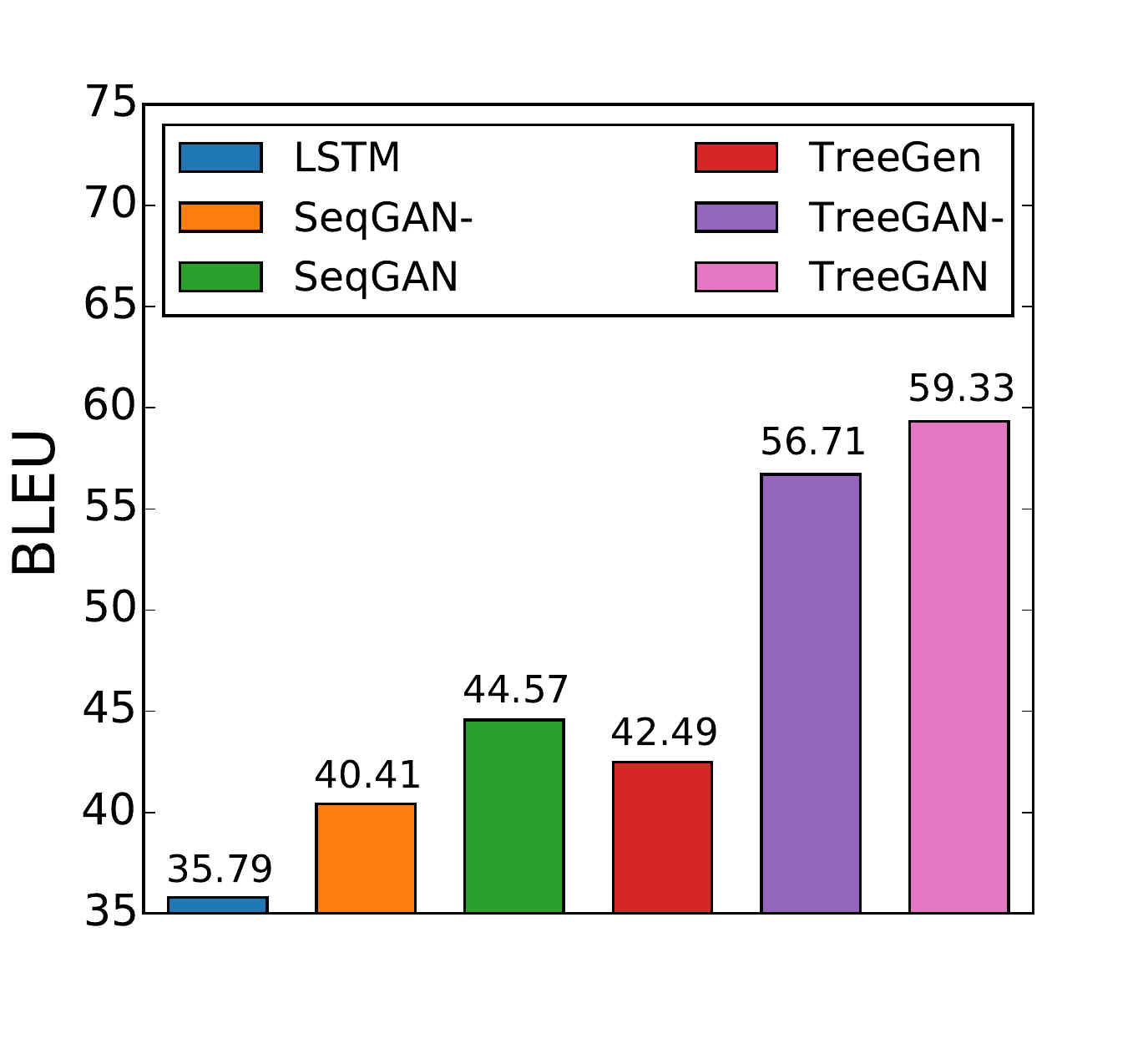}
   \caption{PLD-BLUE} 
  \label{fig:PLD-BLUE}
\end{subfigure}
\begin{subfigure}{.24\textwidth}
  \centering
  \includegraphics[width=1.\linewidth]{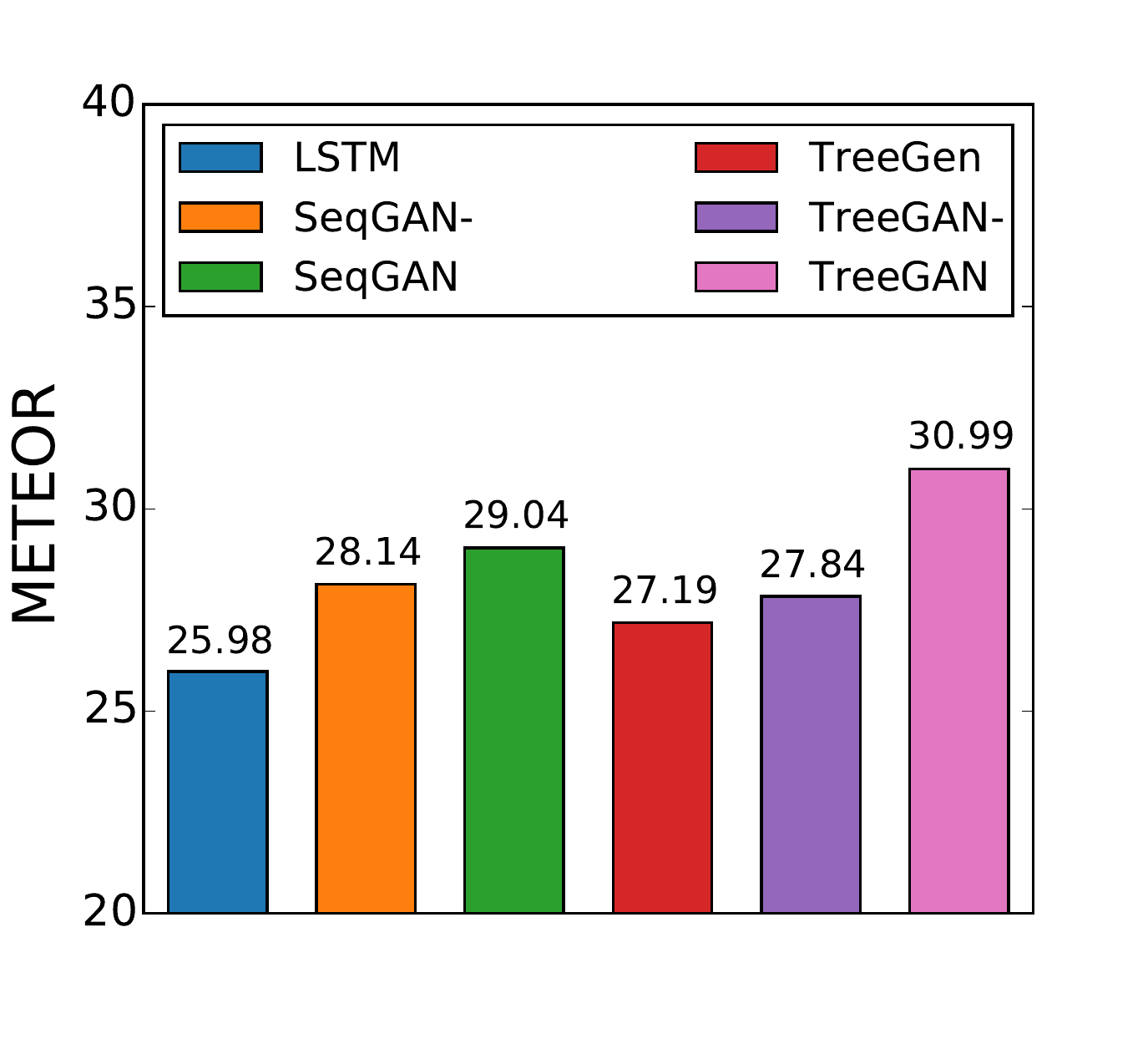}
  \caption{PLD-METEOR}
  \label{fig:PLD-METEOR}
\end{subfigure}
\begin{subfigure}{.24\textwidth}
  \centering
  \includegraphics[width=1.\linewidth]{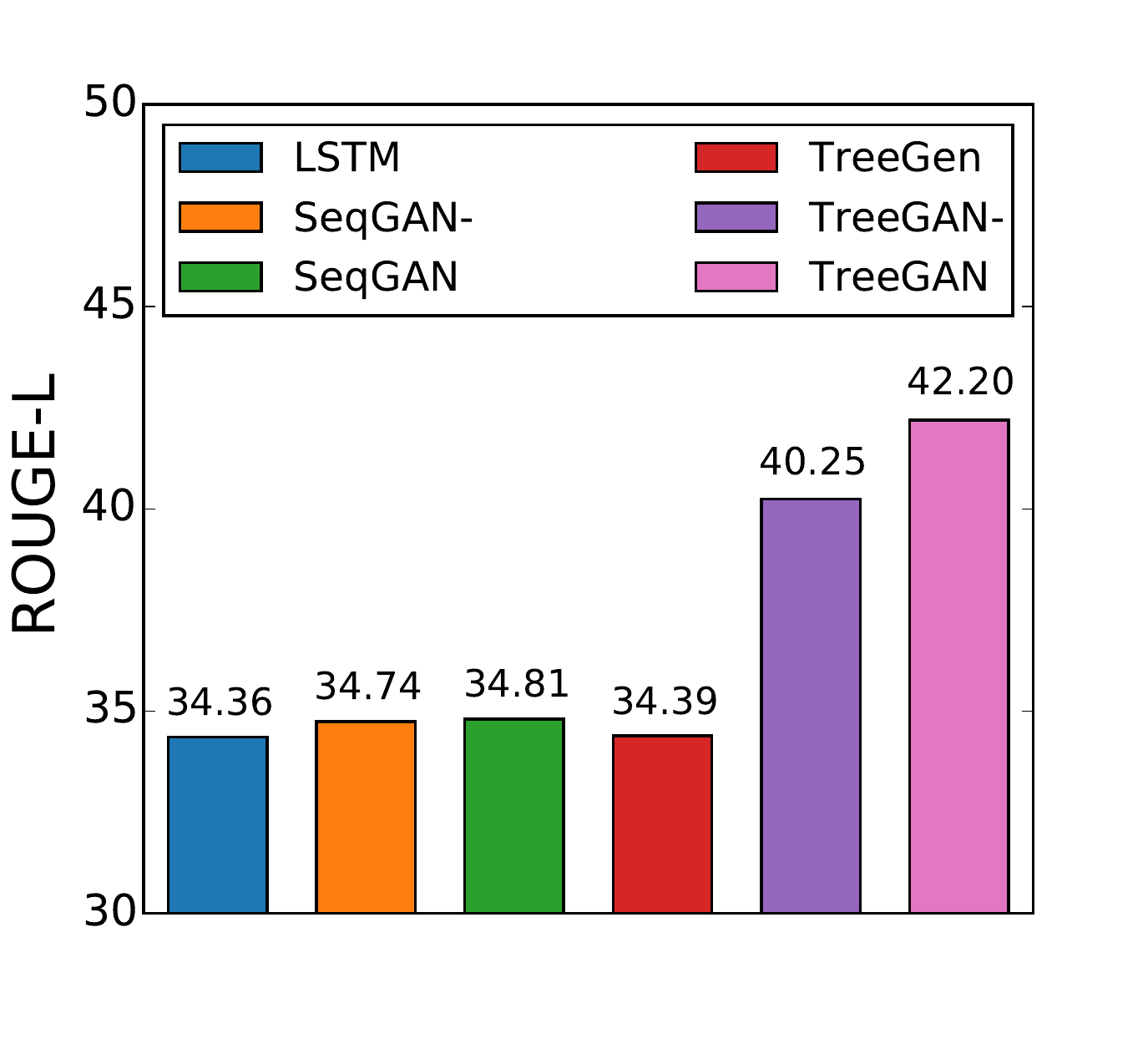}
  \caption{PLD-ROUGE-L} 
  \label{fig:PLD-ROUGEL}
\end{subfigure}
\begin{subfigure}{.24\textwidth}
  \centering
  \includegraphics[width=1.\linewidth]{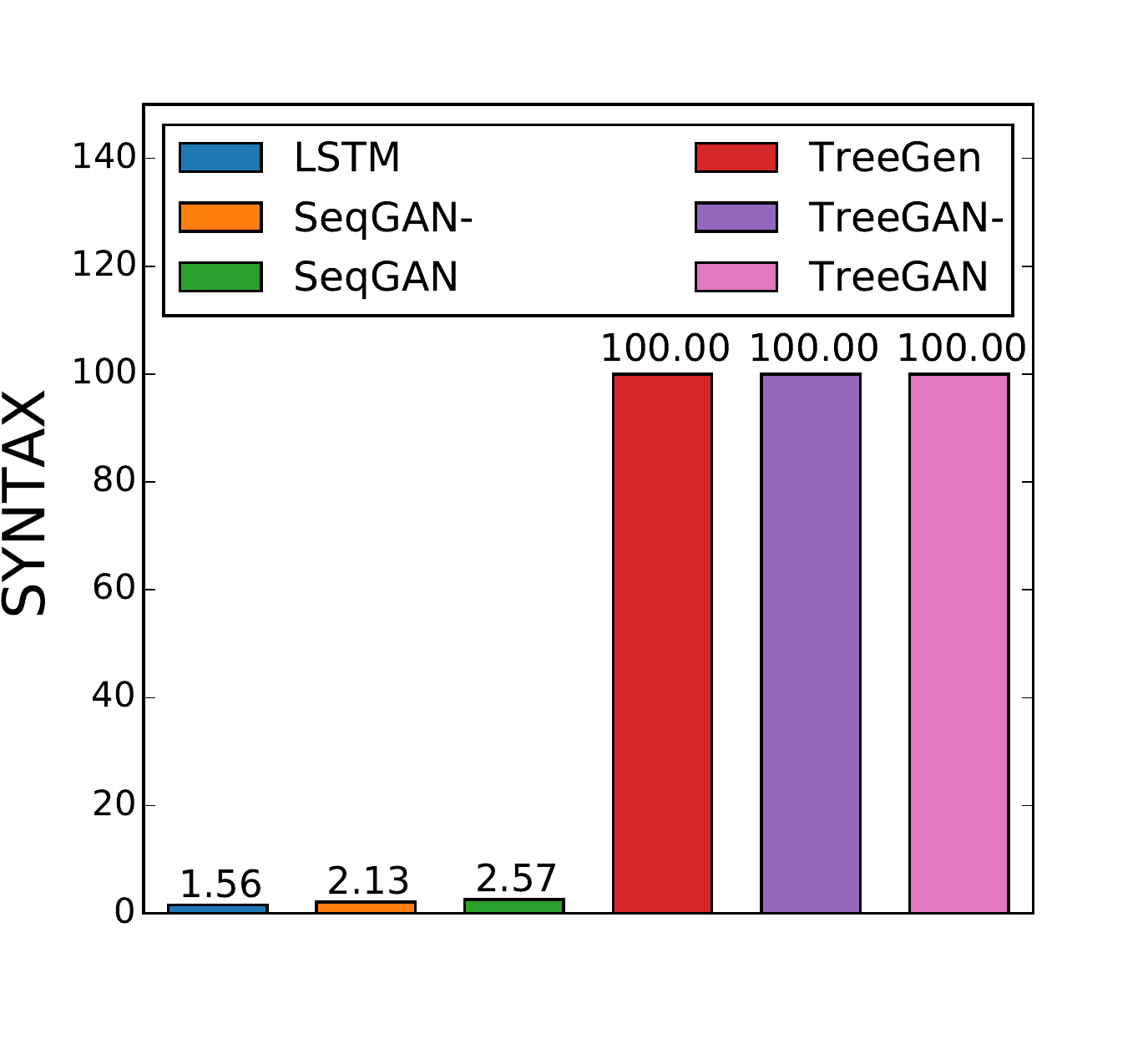}
  \caption{PLD-SYNTAX} 
  \label{fig:treegan}
\end{subfigure}
\caption{Quantitative Evaluation on PLD Dataset.}
\label{fig:result-PLD}
\vspace{-10pt}
\end{figure*} 

\begin{figure*}[t]
\centering
\begin{subfigure}{.19\textwidth}
  \includegraphics[width=1.\linewidth]{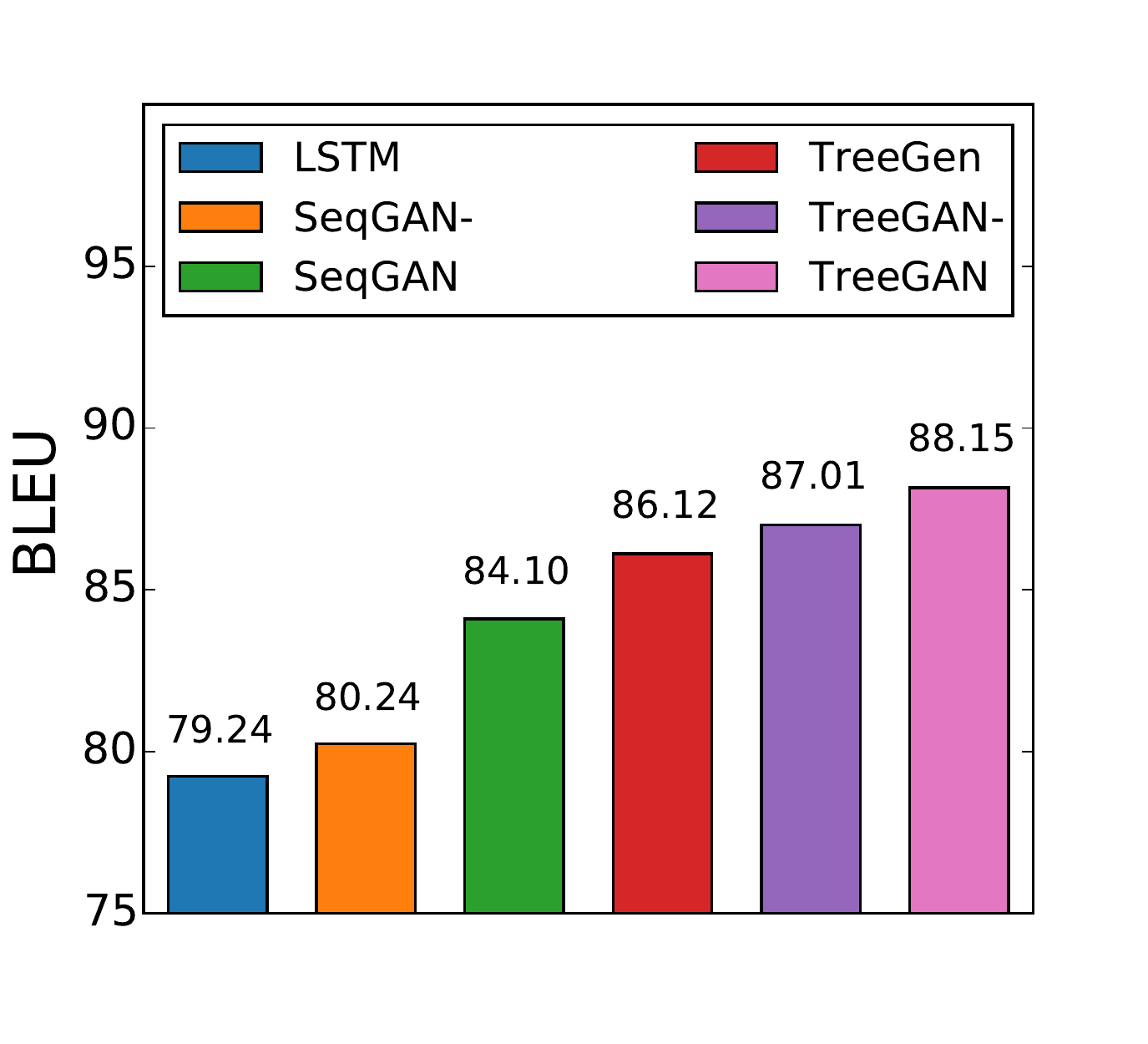}
   \caption{SQL-A-BLUE} 
  \label{fig:dcgan}
\end{subfigure}
\begin{subfigure}{.19\textwidth}
  \centering
  \includegraphics[width=1.\linewidth]{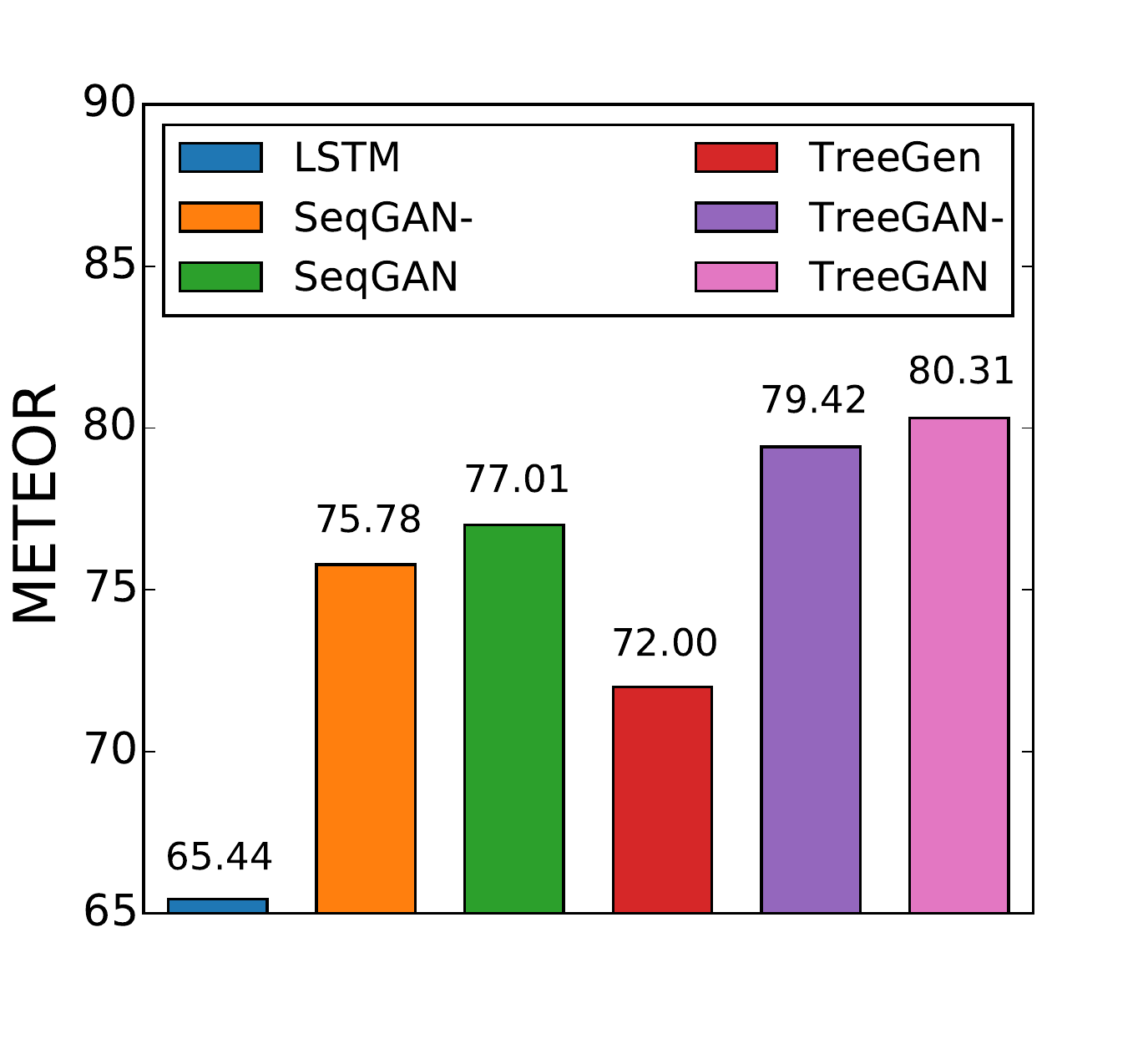}
  \caption{SQL-A-METEOR}
  \label{fig:seqgan}
\end{subfigure}
\begin{subfigure}{.19\textwidth}
  \centering
  \includegraphics[width=1.\linewidth]{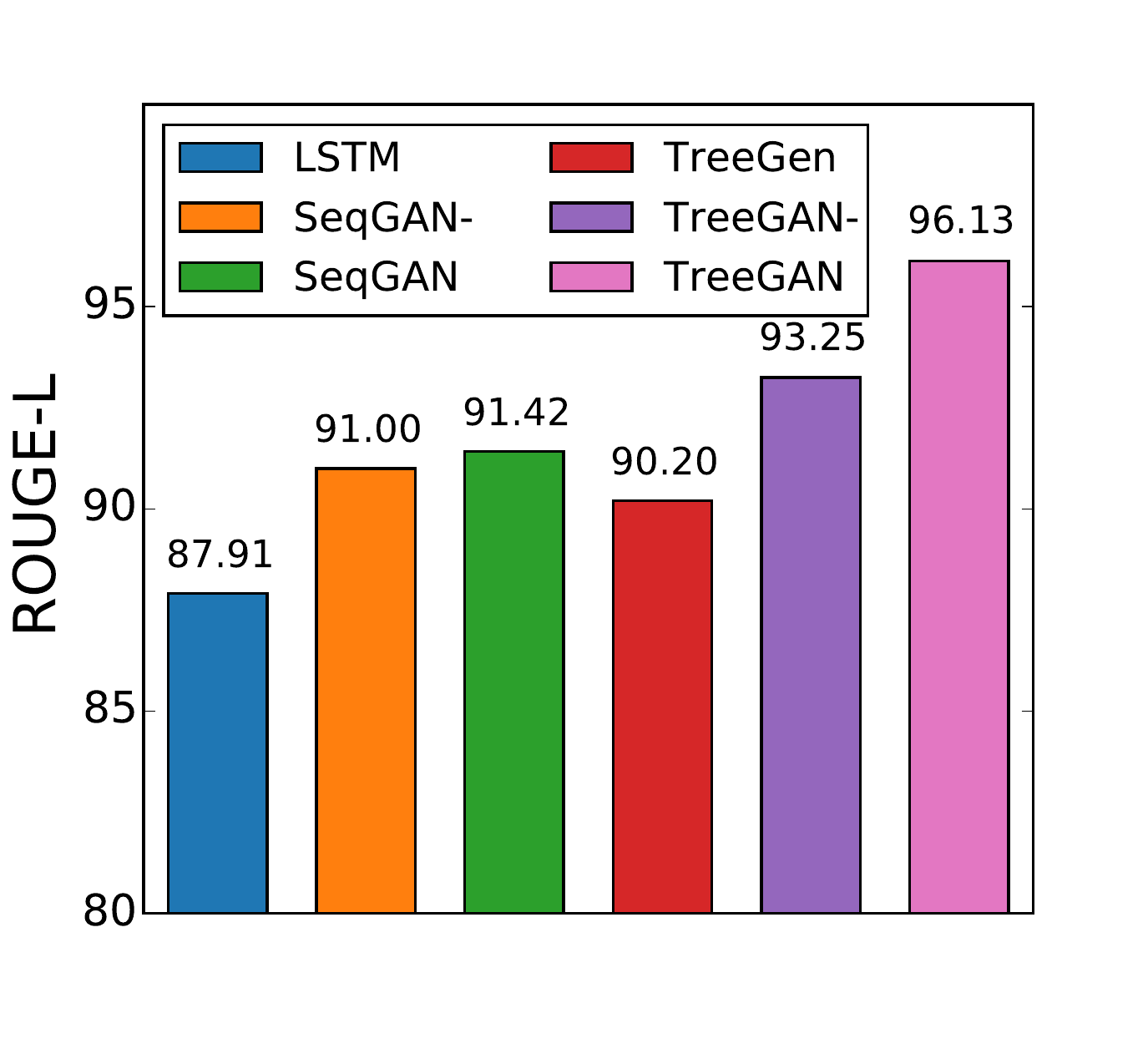}
  \caption{SQL-A-ROUGE-L} 
  \label{fig:treegan}
\end{subfigure}
\begin{subfigure}{.19\textwidth}
  \centering
  \includegraphics[width=1.\linewidth]{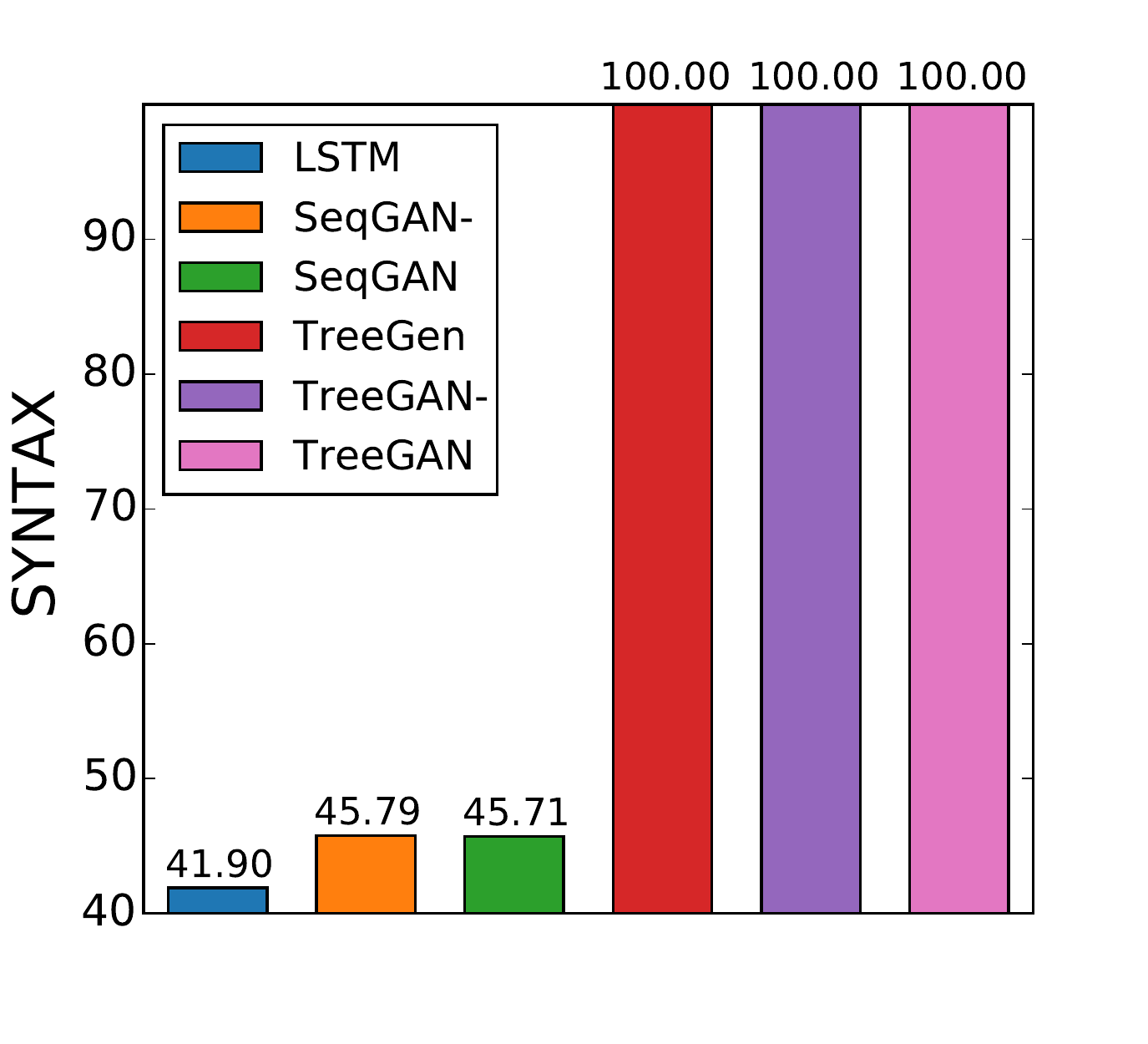}
  \caption{SQL-A-SYNTAX} 
  \label{fig:treegan}
\end{subfigure}
\begin{subfigure}{.19\textwidth}
  \centering
  \includegraphics[width=1.\linewidth]{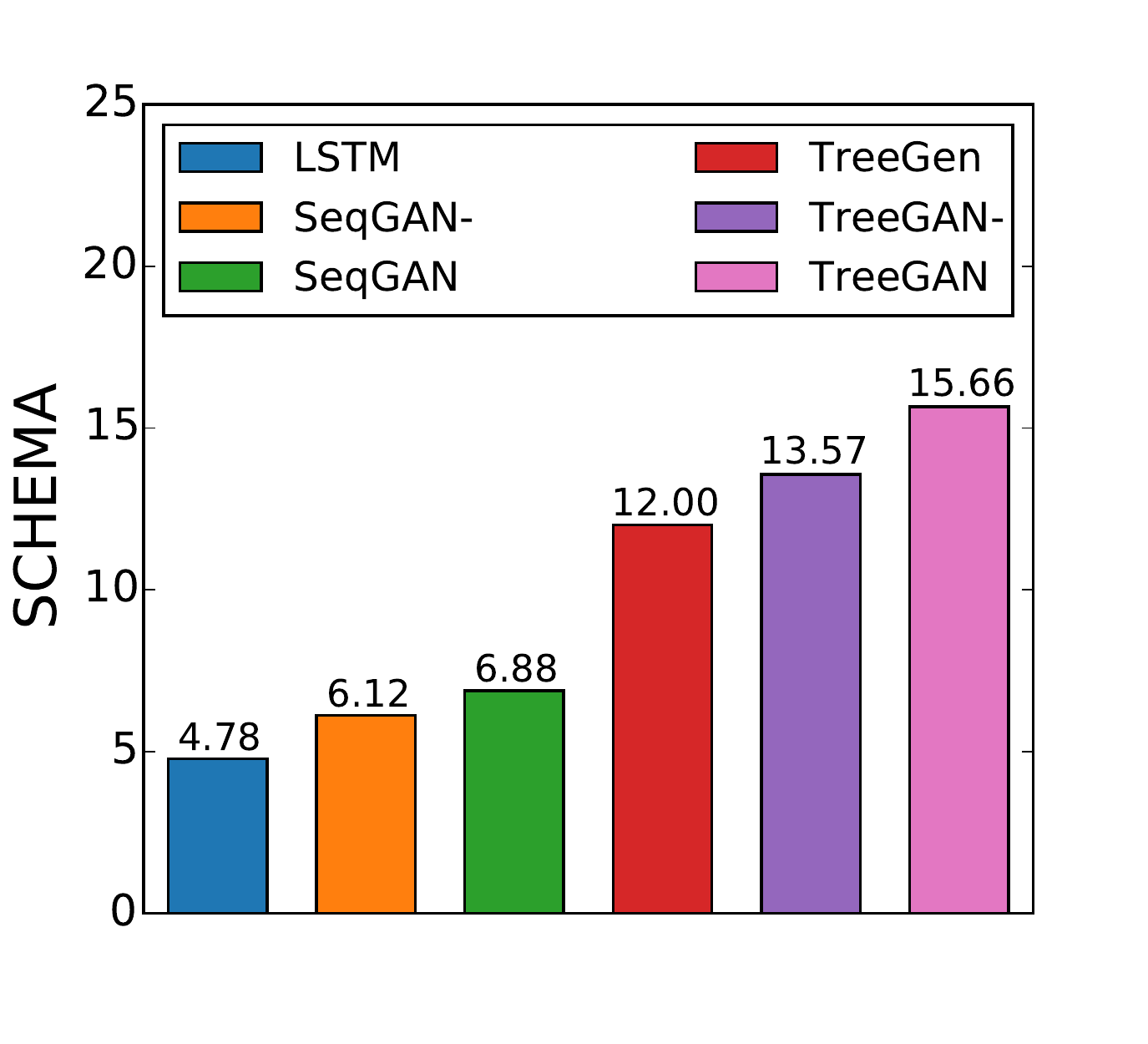}
  \caption{SQL-A-SCHEMA} 
  \label{fig:treegan}
\end{subfigure}
\caption{Quantitative Evaluation on SQL-A.}
\label{fig:result-SQLA}
\end{figure*}

\begin{figure*}[t]
\centering
\begin{subfigure}{.19\textwidth}
  \includegraphics[width=1.\linewidth]{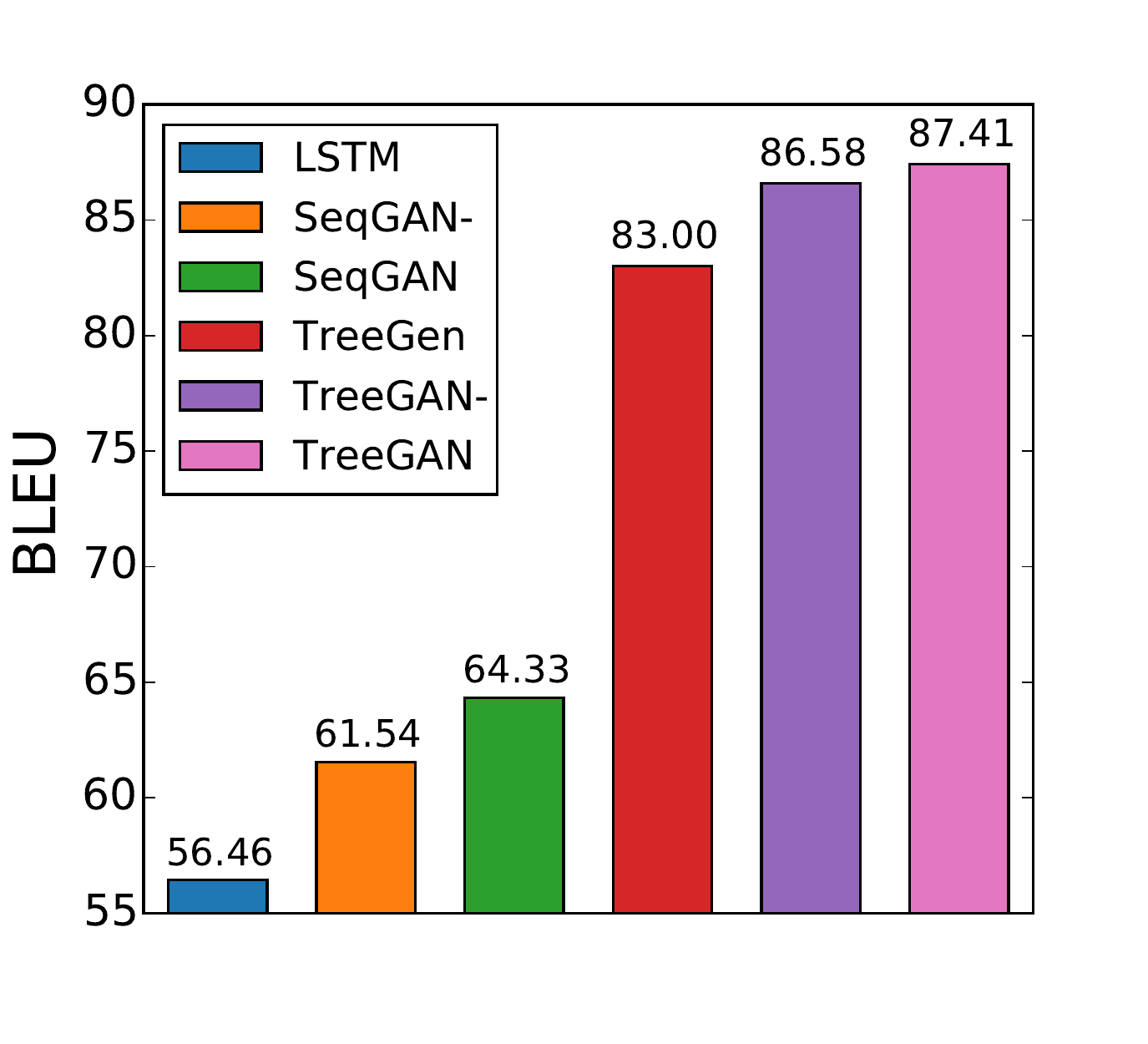}
   \caption{SQL-B-BLUE} 
  \label{fig:dcgan}
\end{subfigure}
\begin{subfigure}{.19\textwidth}
  \centering
  \includegraphics[width=1.\linewidth]{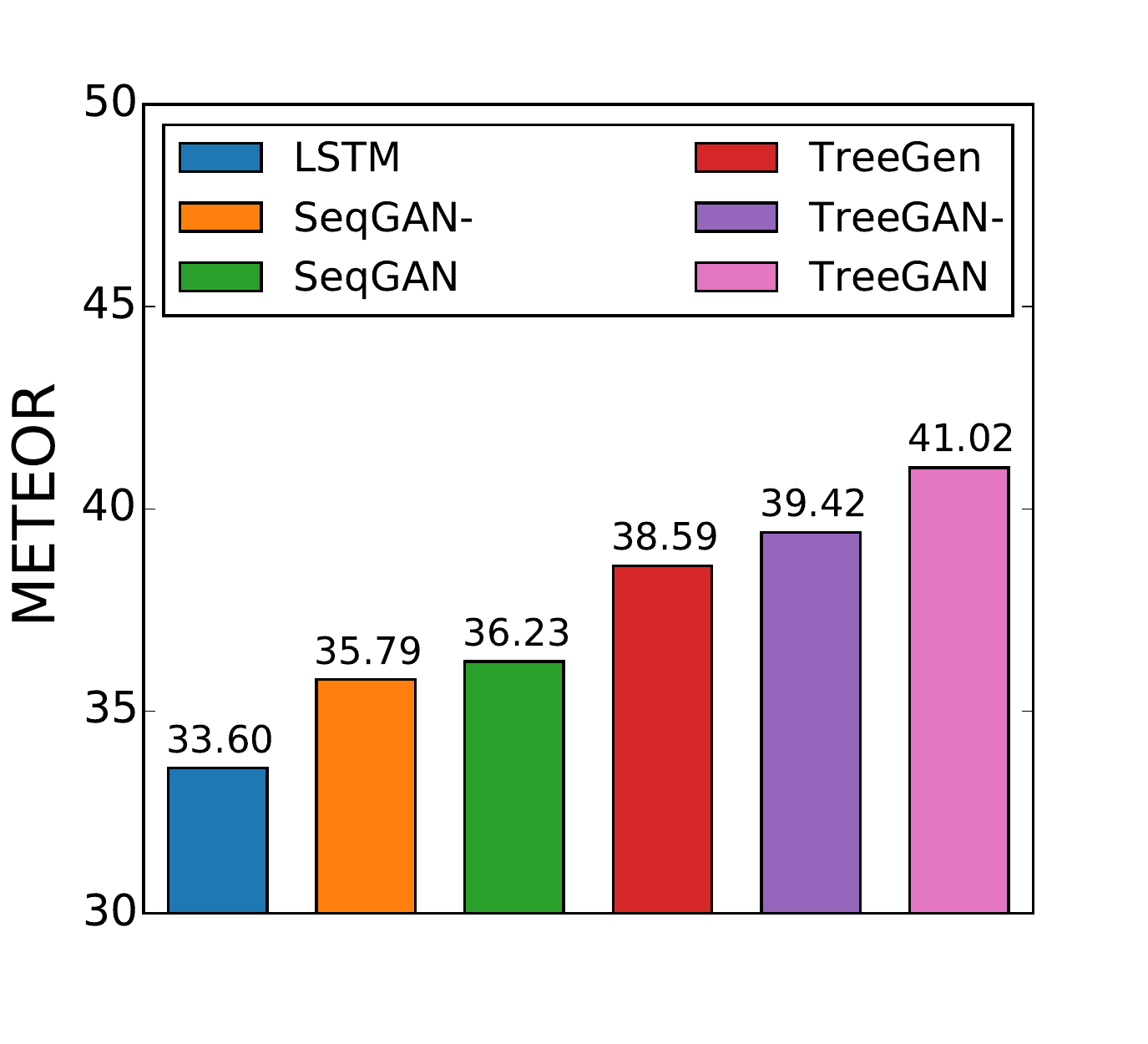}
  \caption{SQL-B-METEOR}
  \label{fig:seqgan}
\end{subfigure}
\begin{subfigure}{.19\textwidth}
  \centering
  \includegraphics[width=1.\linewidth]{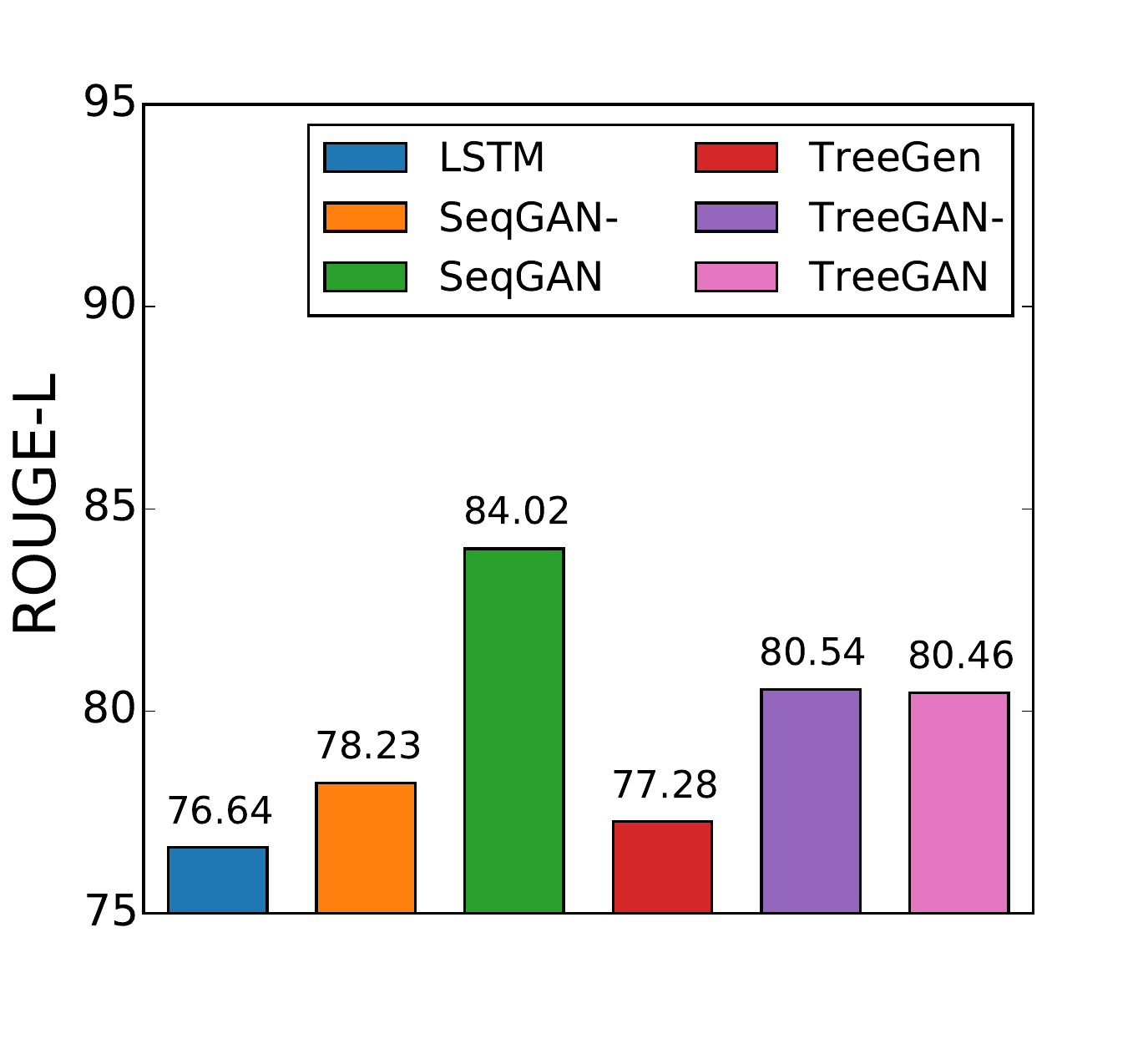}
  \caption{SQL-B-ROUGE-L} 
  \label{fig:treegan}
\end{subfigure}
\begin{subfigure}{.19\textwidth}
  \centering
  \includegraphics[width=1.\linewidth]{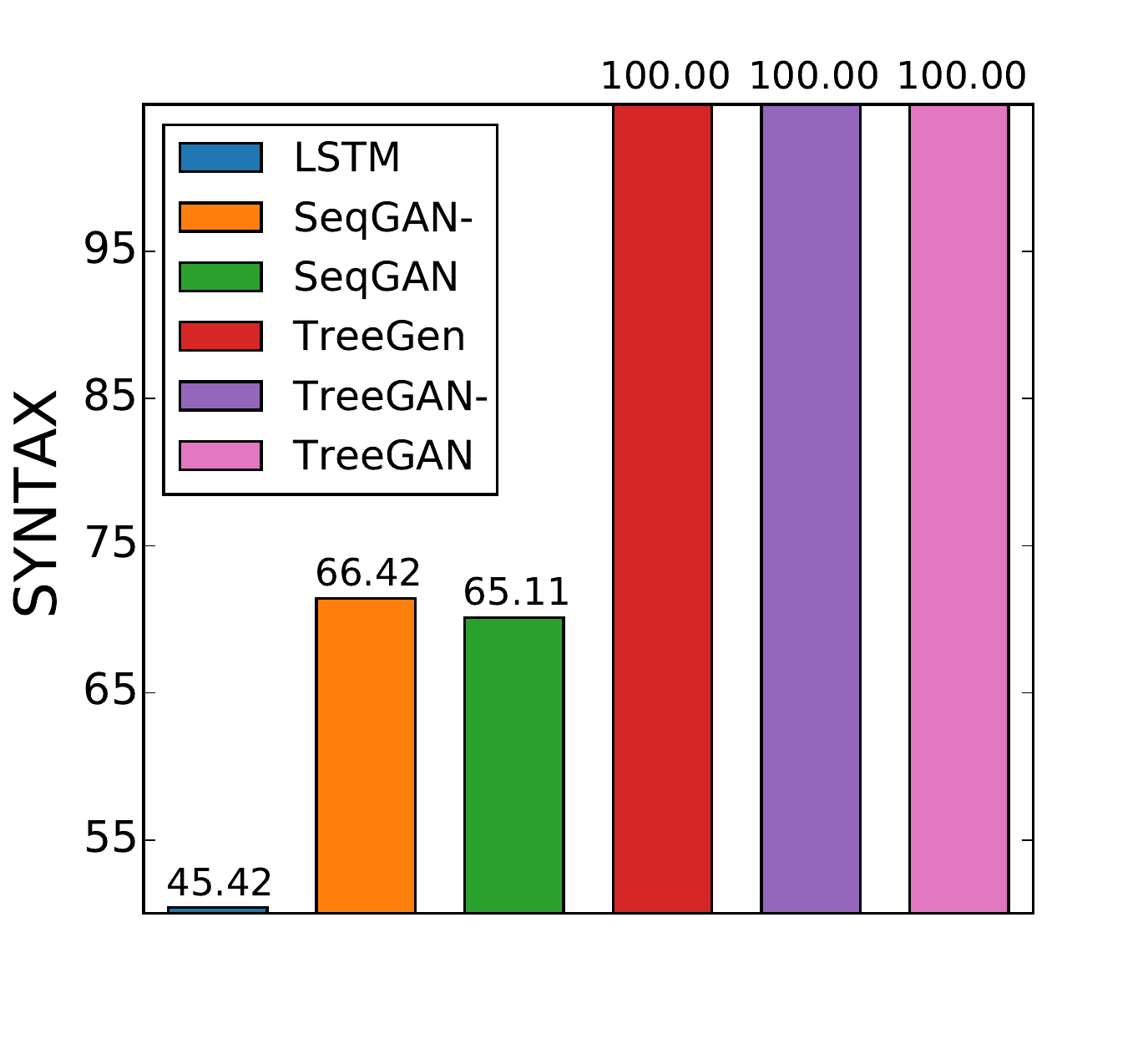}
  \caption{SQL-B-SYNTAX} 
  \label{fig:treegan}
\end{subfigure}
\begin{subfigure}{.19\textwidth}
  \centering
  \includegraphics[width=1.\linewidth]{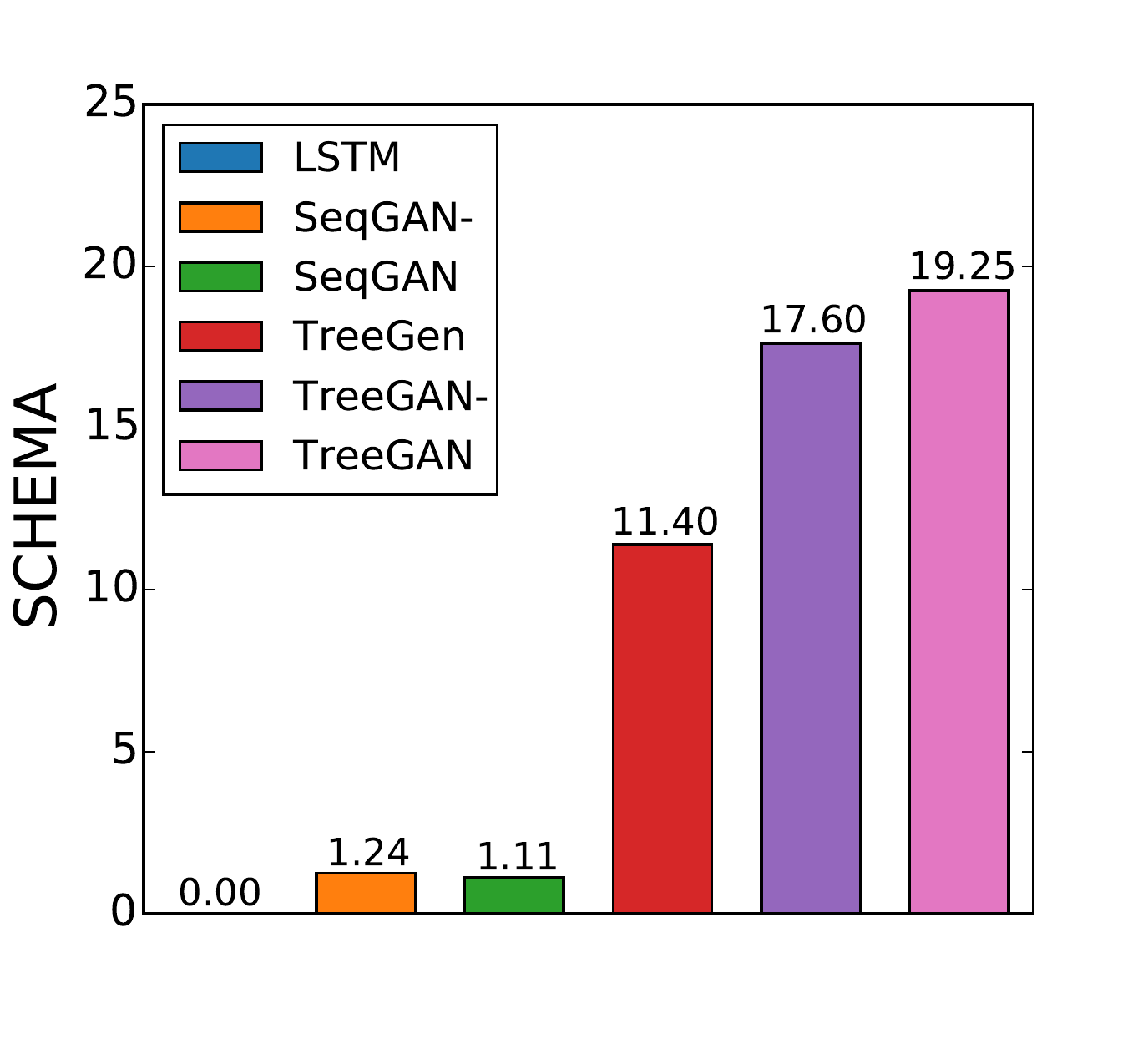}
  \caption{SQL-B-SCHEMA} 
  \label{fig:treegan}
\end{subfigure}
\caption{Quantitative Evaluation on SQL-B.}
\label{fig:result-SQLB}
\end{figure*}

\begin{figure*}[t]
\centering
\begin{subfigure}{.3\textwidth}
  \includegraphics[width=1.\linewidth]{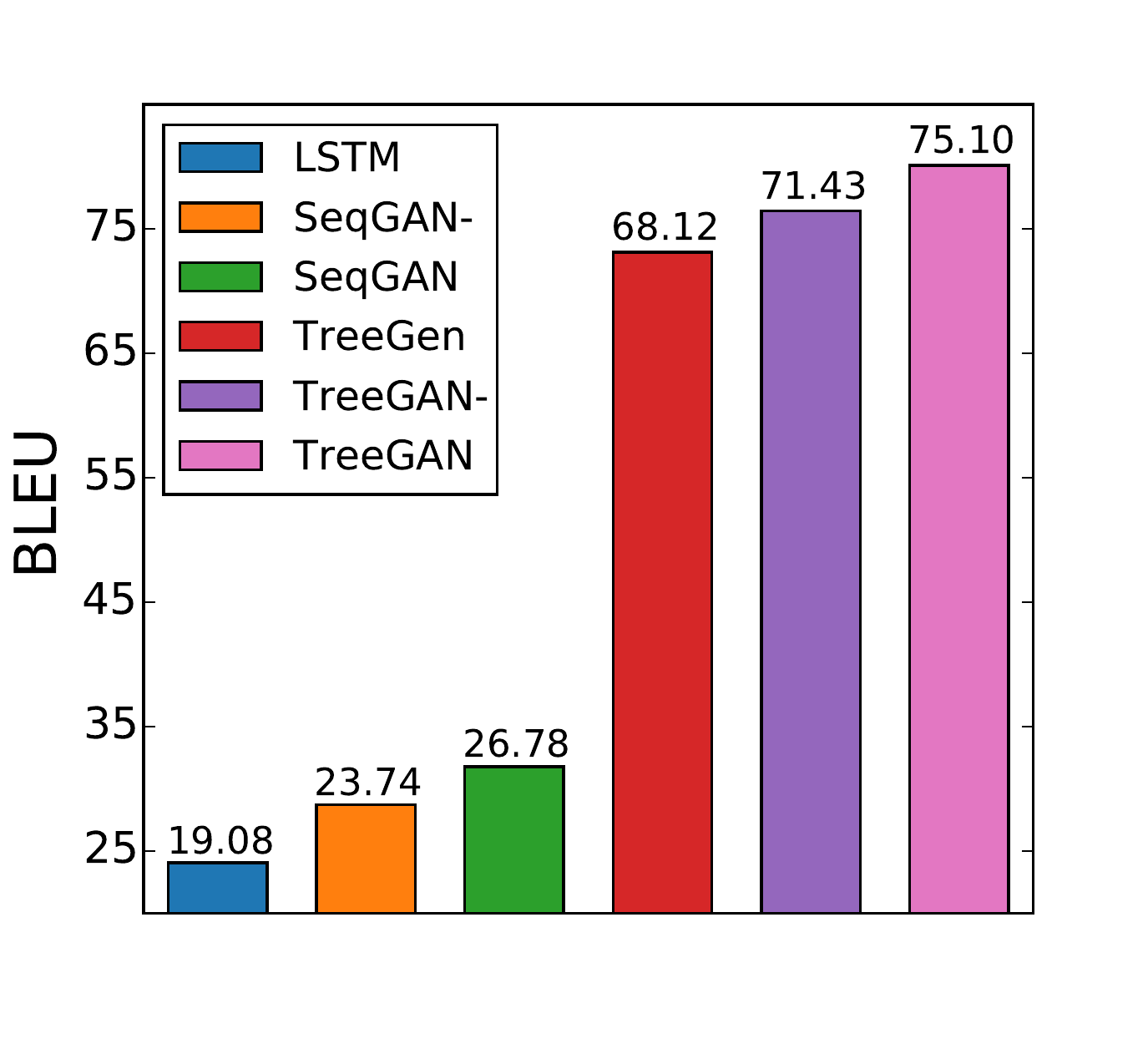}
   \caption{DJANGO-BLUE} 
  \label{fig:dcgan}
\end{subfigure}
\begin{subfigure}{.3\textwidth}
  \centering
  \includegraphics[width=1.\linewidth]{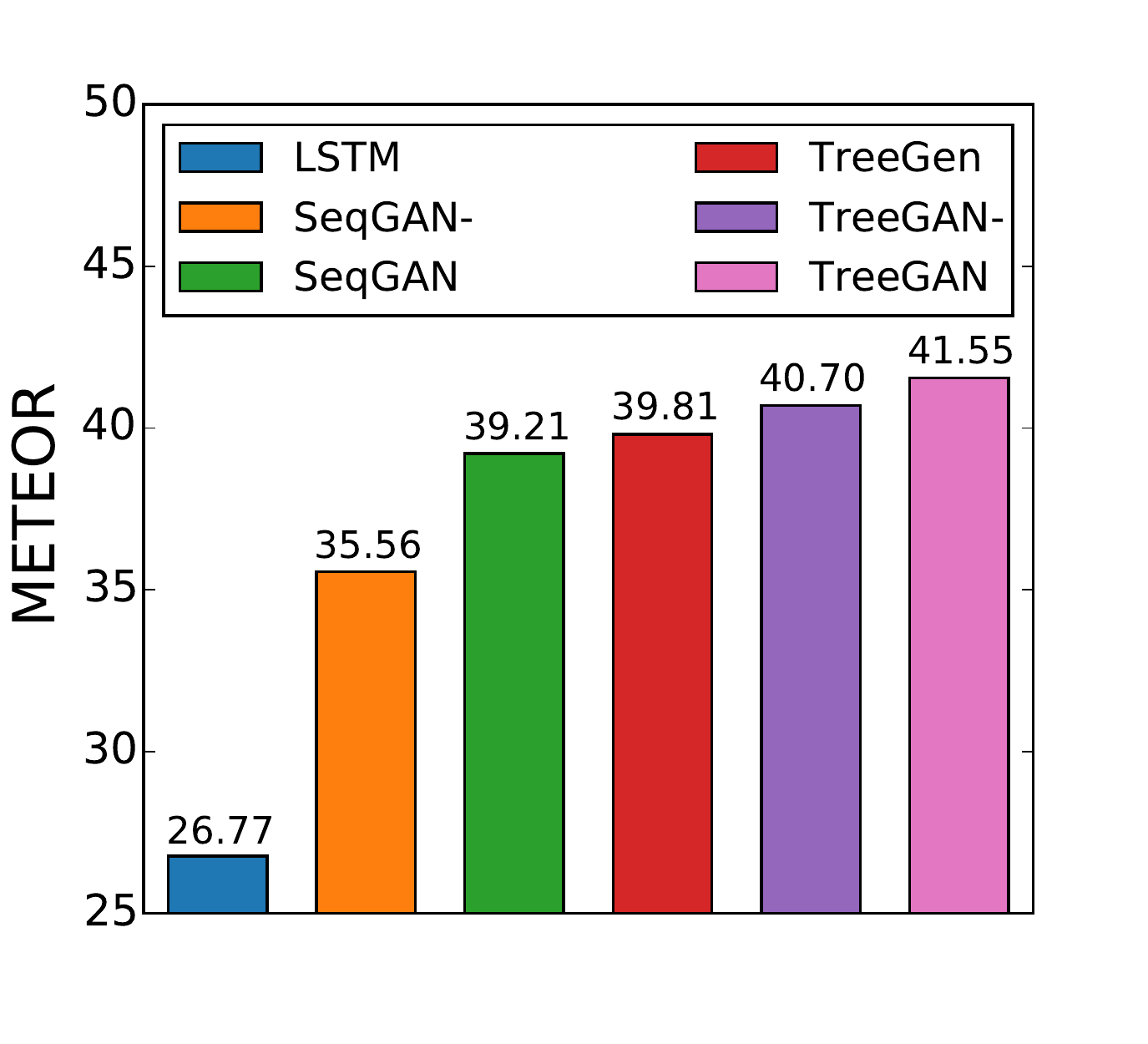}
  \caption{DJANGO-METEOR}
  \label{fig:seqgan}
\end{subfigure}
\begin{subfigure}{.3\textwidth}
  \centering
  \includegraphics[width=1.\linewidth]{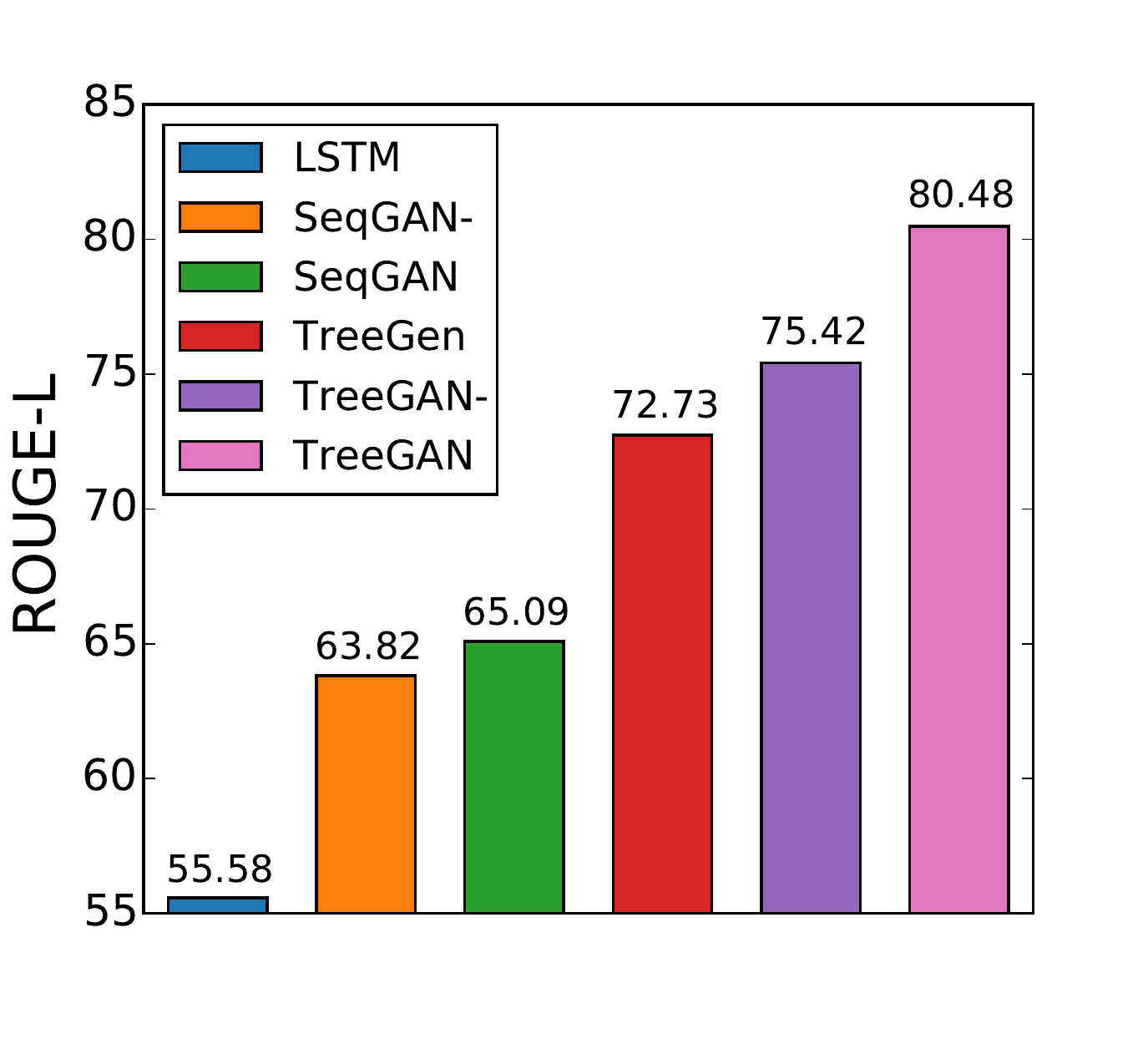}
  \caption{DJANGO-ROUGE-L} 
  \label{fig:treegan}
\end{subfigure}
\caption{Quantitative Evaluation on Django}
\label{fig:result-Django}
\vspace{-10pt}
\end{figure*} 

\subsection{Compared Methods}
We test the following methods to demonstrate the effectiveness of the proposed method.

\noindent $\bullet$ \textbf{TreeGAN} (Our): it uses the tree generator described in Sec.~\ref{sec:method} and the Child-Sum Tree-LSTM as the discriminative model.

\noindent $\bullet$ \textbf{TreeGAN-} (Our): A variation of TreeGAN that uses LSTM as the discriminative model instead of Tree-LSTM.

\noindent $\bullet$ \textbf{TreeGen} \cite{yin2017syntactic}: Tree generator without adversarial training, using MLE training.  

\noindent $\bullet$ \textbf{SeqGAN} \cite{yu2017seqgan}: The original Sequence GAN that proposed for general purpose sequence generation task.

\noindent $\bullet$ \textbf{SeqGAN-} \cite{yu2017seqgan}: A variation of Sequence GAN that uses LSTM as the discriminative model instead of CNN. 

\noindent $\bullet$ \textbf{LSTM} \cite{hochreiter1997long}: LSTM generator employs Maximum Likelihood Estimation as the training strategy.

All compared methods are implemented using PyTorch\footnote{\url{http://pytorch.org}} in Python.
The batch size is set to 64 for all models.

\subsection{Experimental Settings}
For each dataset we used in this section, we first transform each sequence into a sequence of actions that pre-defined in the given syntax.
Note that each sequence of actions represents the yield of syntax parse tree for the corresponding sequence.
Then we randomly select 10\% of data samples to form the test (reference) set, and use the remaining 90\% as the training set. 
For all GAN models include the proposed TreeGAN, we perform $50$ epochs of pre-training before starting the adversarial training.
And the adversarial training last up to $50$ epochs or until the policy gradient loss converges.
We use grid search to find the best hyper-parameter of TreeGAN.
Default hyper-parameter are used for the compared methods unless otherwise stated. 
All the generated trees of TreeGAN are translated into sequence for evaluation purpose.
We report the evaluation scores based on the generations of trained generative net against the samples in the test (reference) set. 
We used the number of generations produced by the trained generator to be the same as the size of test set in each dataset.

\subsection{Evaluation Metrics} 
We include commonly used metrics such as \texttt{BLEU}-3 \cite{papineni2002bleu}, \texttt{METEOR} score \cite{denkowski2014meteor} and \texttt{ROUGE-L} score \cite{lin2004rouge}. 
Since neither of these metrics is designed to measure how well the generated sequences fit the target grammar, we propose two additional metrics to evaluate them. 
The first one measures the percentage of the generated sequences that are grammatically correct (labeled as \texttt{SYNTAX}).
For SQL generation tasks, we additionally report the percentage of generated sequences which obey the schema (labeled as \texttt{SCHEMA}, evaluate the correctness of entity and relation for the generated SQL). 

\subsection{Quantitative Results}
Figure~\ref{fig:result-PLD}, Figure~\ref{fig:result-SQLA} and Figure~\ref{fig:result-SQLB} show the quantitative results on synthetic dataset PLD, SQL-A, SQL-B respectively.

To answer the \textbf{RQ1}, we demonstrate the \texttt{SYNTAX} scores in Figure~\ref{fig:result-PLD}(d), Figure~\ref{fig:result-SQLA}(d) and Figure~\ref{fig:result-SQLB}(d), in which we observe that the tree-based frameworks, including the proposed TreeGAN, achieve 100\% syntax correctness regards the pre-defined grammar while other baselines perform badly in terms of this syntax correctness.
This results show that the proposed TreeGAN could fully capture the given syntax information and generate grammatically correct sequence.

As to the \textbf{RQ2}, we could read Figure~\ref{fig:result-SQLA}(e) and Figure~\ref{fig:result-SQLB}(e).
We discover that even though without explicit input of schema, the proposed TreeGAN has higher chance for capturing the underlying semantic pattern, given at least $3.66\%$ and $7.85\%$ improvement on \texttt{SCHEMA} in SQL-A and in SQL-B respectively.   

More generally, we use three popular NLP metrics to evaluate the quality of the generated sequences, which are presented in Figure~\ref{fig:result-PLD}(a)-(c), Figure~\ref{fig:result-SQLA}(a)-(c) and Figure~\ref{fig:result-SQLB}(a)-(c).
From these figures we can clearly see the superiority of TreeGAN, who consistently outperforms the compared methods in terms of the \texttt{BLEU}, \texttt{METEOR} and \texttt{ROUGE-L} (except the case of \texttt{ROUGE-L} in SQL-B, where TreeGAN still obtain competitive results).
These results should clearly answer the \textbf{RQ3} we raised earlier, the quality of sequences generated by TreeGAN is better than the generations of compared methods.

\subsection{Qualitative Results}
Table~\ref{tab:syn_text} samples several generations on SQL-B dataset for qualitative evaluation.
We mainly compare the generations of TextGAN with SeqGAN to demonstrate the advantage of employing tree structure generator and discriminator in GANs on sequence.
Consistent with the results we have seen in quantitative evaluation, SeqGAN's generations could not perfectly follow the underlying grammar and exhibit syntax errors (highlighted in {\color{red}\textbf{red}}).
As shown in Table~\ref{tab:syn_text}, the generation \circled{4} of SeqGAN mistakenly applies `count' aggregation on a numerical value and does not close the `from' clause correctly.
Similar syntax errors can also be observed in the generation \circled{5} and \circled{6} by SeqGAN.
Meanwhile, TreeGAN incorporates the pre-defined grammar, and all its generations are valid.
We randomly select some examples in Table~\ref{tab:syn_text}.
The generation  \circled{7} and \circled{8} mimic the ground truth \circled{1} and \circled{2} well and capture the underlying schema correctly.
Generation \circled{9} resembles ground truth \circled{3} and extend it with an extra `where' clause.
These observations contribute to \textbf{RQ1} and \textbf{RQ3}, and re-confirm the answers we obtained in the analysis of quantitative results.

\begin{table}[!t] 
	\small
	\centering
    \caption{SQL query generation in SQL-B. Syntax errors are highlighted in {\color{red}\textbf{red}} color.}
    \label{tab:syn_text}
    \begin{tabular}{l}
    \toprule
    \textbf{Real SQL Queries} \\
    \circled{1} select count(authenticated) from America where alight$>$3;\\
    \circled{2} select driftpin, min(deject) from Danmark where driftpin$=$16;\\
    \circled{3} select hedy from Hungary;\vspace{2pt}\\
    \toprule
    \textbf{Queries generated by SeqGAN}\cite{yu2017seqgan} \\
    \circled{4} select \textbf{{\color{red}count(17)}}, min(acoustically) \textbf{{\color{red}from;}}\\
    \circled{5} select max(cookstove), gainfully, \textbf{{\color{red} min ())}}, \\
    \hspace{3mm} min(buttonhole) from America;\\
    \circled{6} select aalesund from Brazil where hanuman \\
    \hspace{3mm}\textbf{{\color{red} acoustically Hungary}};\\
    \toprule
    \textbf{Queries generated by TreeGAN} (our method)\\ 
    \circled{7} select min(jacarta) from Jamaica;\\
    \circled{8} select min(endogenous) from Brazil where epigraphical=1;\\
    \circled{9} select hedy from Hungary where deject!=2;\\
    \bottomrule
    \end{tabular}
\end{table}

\section{Experiments on Real Data}
\label{sec:real-data}

To better demonstrate the superiority of the proposed TreeGAN, we also perform experiments on real dataset.
We try to additionally answer the following research questions through this section.
\begin{itemize}
\item \textbf{RQ4}: Does TreeGAN 	achieve similar performance on real dataset with more complex syntax as in synthetic dataset?
\item \textbf{RQ5}: What is the limitation of TreeGAN on sequence generation?
\end{itemize}

\subsection{Dataset}
We test our proposed model on the python code dataset \cite{oda2015learning} from django\footnote{\url{https://www.djangoproject.com/}} project. 
It is a collection of lines of python code, and each performs a functional task.
We use Python AST package and Astor package\footnote{\url{http://astor.readthedocs.io/en/latest/}} to construct and parse the AST corresponds to each line of code in the dataset.
The code in Django dataset is diverse and spanning a wide variety of real-world use cases such as I/O operations, exception handling, and mathematical computation.
We follow the same setting and same evaluation metrics (\texttt{SYNTAX} is not reported due to the freeness of Python grammar) as in the previous section. 
  
 \subsection{Quantitative Results}
Figure~\ref{fig:result-Django} shows the results quantitative evaluation on Django dataset, from which we discover TreeGAN achieves 6.82\% improvement against TreeGen and 18.14\% improvement against SeqGAN in terms of \texttt{BLEU} score. 
As to the METEOR score, TreeGAN improves the performance 1.75\% against TreeGen and 2.34\% against SeqGAN.
We also discover obvious improvement has been made by TreeGAN against TreeGen and SeqGAN in terms of \texttt{ROUGE-L}.
Hence, TreeGAN exhibits similar advantages on the real data as in the synthetic study, which gives a positive answer towards \textbf{RQ4}.

\subsection{Qualitative Results}
Table~\ref{tab:real_text} shows generations from TreeGAN and SeqGAN on Django dataset.
Similar to the results obtained in the synthetic study, we found although SeqGAN could mimic the real Python code, it exhibits several types of syntax errors (highlighted in {\color{red}\textbf{red}}). 
Generation \circled{4} indicates SeqGAN sometimes could not correctly fill the function arguments, generation \circled{5} exhibits a misunderstanding of import statement, while generation \circled{6} demonstrates SeqGAN has difficulty in pairing the parentheses. 
Meanwhile, generation \circled{7} and \circled{8} show the capability of TreeGAN on learning the usage of assignment statement, function call, conditional statement,{\etc}
These observations indicate that on code generation tasks, TreeGAN could effectively plug in complex grammatical rules and generate valid code snippets, which re-confirm the answer we obtained for \textbf{RQ4}.

We also discuss the limitation of TreeGAN (\textbf{RQ5}), which could shed a light on future extension.     
From generation \circled{9}, we can see TreeGAN has difficulty in understanding the concept of inheritance and the member function, where the parent class of `META' does not have the member function called `new\_file()', the call is invalid and causes a running-time error.
There are about 3.7\% of generations by TreeGAN exhibit the similar semantic error in our experiments.
It is not difficult to identify that this semantic error in Python is the counterpart of the schema error in SQL.  
It is possible for our model to learn these semantic pattern from the data, but it may need a better way to guide the learning process for fully capturing the semantic, which could be a future direction for this work.

\subsection{Final Remarks}
Through the analysis of synthetic study and experiments on real data, we can finally answer the following two research questions to summarize our experiments:
\begin{itemize}
\item \textbf{RQ6}: Is TreeGAN a better \textit{GAN model} on sequence generation?
\item \textbf{RQ7}:	Is TreeGAN a better \textit{syntax-aware model} on sequence generation?
\end{itemize}

The quantitative and qualitative comparisons between TreeGAN and SeqGAN (and its variation) should support a positive answer towards the \textbf{RQ6}. 
As to the \textbf{RQ7}, we can conclude that employing GAN improves the generation quality by comparing TreeGAN with TreeGen (as shown in synthetic study and real data experiments).
By convincing TreeGAN is a better GAN model and a better syntax-aware model on sequence generation, we justify that both \textit{syntax-aware} and \textit{GAN} are indispensable components toward a more useable sequence generation.

\begin{table}[t] 
	\small
	\centering
    \caption{Python code generation in Django. Syntax errors are highlighted in {\color{red} \textbf{red}} color.}
    \label{tab:real_text}
    \begin{tabular}{l}
    \toprule
    \textbf{Real Python Code} \\
    \circled{1} f.write(pickle.dumps(expiry,-1))\\
    \circled{2} db = router.db\_for\_read(self.cache\_model\_class )\\
    \circled{3} {if connections[db].features.needs\_datetime\_string\_cast and not } \\ 
    \hspace{3mm}{ isinstance(expires, datetime)}\\
    \toprule
    \textbf{Code generated by SeqGAN}~\cite{yu2017seqgan} \\
    \circled{4} name=self.\_save(\textbf{{\color{red},}} name, content, self)\\
    \circled{5} from django.ImproperlyConfigured \textbf{{\color{red} import 0}}\\
    \circled{6} return urljoin(self.base\_url, filepath\_to\_uri\textbf{{\color{red}()))}}\\
    \toprule
    \textbf{Code generated by TreeGAN} (our method) \\ 
    \circled{7} {table = connections[db].ops.quote\_name(self.\_table) }\\
    \circled{8} if exp is None or exp$>$time.time()\\
    \circled{9} super(META, self).new\_file(file\_name, *args)\\
    \bottomrule
    \end{tabular}
\end{table}

\section{Related Work}
\label{sec:rel}

Our work is related to both syntax aware sequence generation and generative adversarial networks (GANs), we briefly discuss them respectively in this section.

\subsection{Syntax Aware Sequence Generation}
Most works on this line require descriptive input such as text specification \cite{lei2013natural, ling2016latent, balog2016deepcoder,yin2017syntactic}, and their overall goal is to generate code that performs the corresponding task(s) described in the input text.
Our proposed model is different from these existing models in several aspects:
(1) Our model does not require any descriptive text as the input.
(2) Our model employs a GAN training framework to improve the generation quality.
(3) Our model targets at generating arbitrary sequences that follow the pre-defined syntax and resemble the real sequences.
Some other methods on code generation focus on specific languages \cite{raza2015compositional, parisotto2016neuro}, but our model is generalized to fit any context-free grammar.
Besides, there are several probabilistic generation models \cite{maddison2014structured, nguyen2013statistical}, which are mainly based on Bayesian estimation while our work is based on neural networks.

\subsection{Generative Adversarial Networks (GANs)}
Figure~\ref{fig:compare} illustrates the comparison between TreeGAN and the related GAN generation methods.
GAN was first proposed in \cite{goodfellow2014generative}, and it exhibits superb performance on image generation \cite{isola2017image, denton2015deep} and image synthesis \cite{reed2016generative, zhang2017stackgan}.
Later on, \cite{yu2017seqgan} studied GAN on sequence generation using policy gradient and Monte Carlo search.
\cite{zhang2017adversarial} alternatively employs feature matching to perform the similar task.
\cite{hu2017toward} additionally consider adding control on the sentiment and tenses of the generated text.
However, neither of these works consider the existing grammar or syntax of the target language, and the generated text may exhibit syntax errors.
Our work makes the first effort to incorporate the grammatical knowledge into GAN model on sequence and text generation.

\section{Conclusion and Future Work}
\label{sec:con}
We proposed a syntax-aware GAN model called TreeGAN for sequence generation. 
We transform the problem into parse tree generation to incorporate the rich grammar information, and both the generator and discriminator are well-tailored to encode the syntax properly.
The experiments on both synthetic datasets and real-world datasets demonstrate that TreeGAN is a promising adversarial learning framework for syntax-aware sequence generation.

We plan to extend this work in two directions in the future.
The first extension will focus on incorporating pre-defined schema information ({\eg} SQL schema) into the GAN model, which allows the generator to fully compatible with the finer level semantics of the target formal language and extend the incorporated grammar to the scope of context sensitive grammar (CSG).
The second extension will consider a topic-wise TreeGAN, which could not only generate valid sequences under the given grammar, but ensure all the generated sequences are describing the given target topic.
With the proposed TreeGAN and these potential extensions, we could make GANs a more practicable and versatile tool for automatically composing sequence.

{\small
\balance
\bibliographystyle{IEEEtran}
\bibliography{header,reference}}
\end{document}